\documentclass[11pt]{article}

\usepackage[final]{acl}

\usepackage{times}
\usepackage{latexsym}

\usepackage[T1]{fontenc}

\usepackage[utf8]{inputenc}

\usepackage{microtype}

\usepackage{inconsolata}

\usepackage{graphicx}

\usepackage{pdfpages}
\usepackage{pifont}
\usepackage{enumitem}
\usepackage{booktabs}
\usepackage{multirow}
\usepackage{colortbl}
\usepackage[table,dvipsnames]{xcolor}
\usepackage{makecell}
\usepackage{arydshln}
\usepackage{amssymb}
\usepackage{xspace}
\usepackage[most]{tcolorbox}
\usepackage{multicol}

\definecolor{my_green}{RGB}{40,154,121}
\definecolor{my_yellow}{RGB}{255,165,0}
\definecolor{my_red}{RGB}{176,46,46}
\definecolor{SPUR1}{RGB}{0,80,160}
\definecolor{SPUR2}{RGB}{40,100,180}
\definecolor{SPUR3}{RGB}{80,120,200}
\definecolor{SPUR4}{RGB}{120,140,220}

\newcommand{\ours}{SPUR\xspace}

\newcommand{\eg}{\hbox{e.g.,}\xspace}
\newcommand{\ie}{\hbox{i.e.,}\xspace}
\newcommand{\best}[1]{\cellcolor{ForestGreen!30}\textbf{#1}}
\newcommand{\second}[1]{\cellcolor{blue!15}\underline{#1}}

\definecolor{ForestGreen}{RGB}{34,139,34}
\newcommand{\tablestyle}[2]{\setlength{\tabcolsep}{#1}\renewcommand{\arraystretch}{#2}}

\usepackage{adjustbox}  
\usepackage{float}      

\definecolor{mask_red}{RGB}{196, 38, 44}      
\definecolor{mask_yellow}{RGB}{194, 160, 4}   
\definecolor{mask_blue}{RGB}{0, 133, 202}     
\definecolor{LastBule}{RGB}{192,211,235}
\definecolor{LastYellow}{RGB}{244,245,208}
\definecolor{myblue}{RGB}{39,108,191}
\definecolor{backblue}{RGB}{244,247,251}
\definecolor{mypink}{RGB}{244,231,250}
\definecolor{myorange}{RGB}{255,236,231}
\definecolor{bggreen}{RGB}{224,236,233}
\definecolor{mypurple}{RGB}{217,205,246}
\usepackage{hyperref}

\usepackage{fontawesome5} 

\newcommand{\homepage}{\raisebox{-1.5pt}{\includegraphics[height=1.2em]{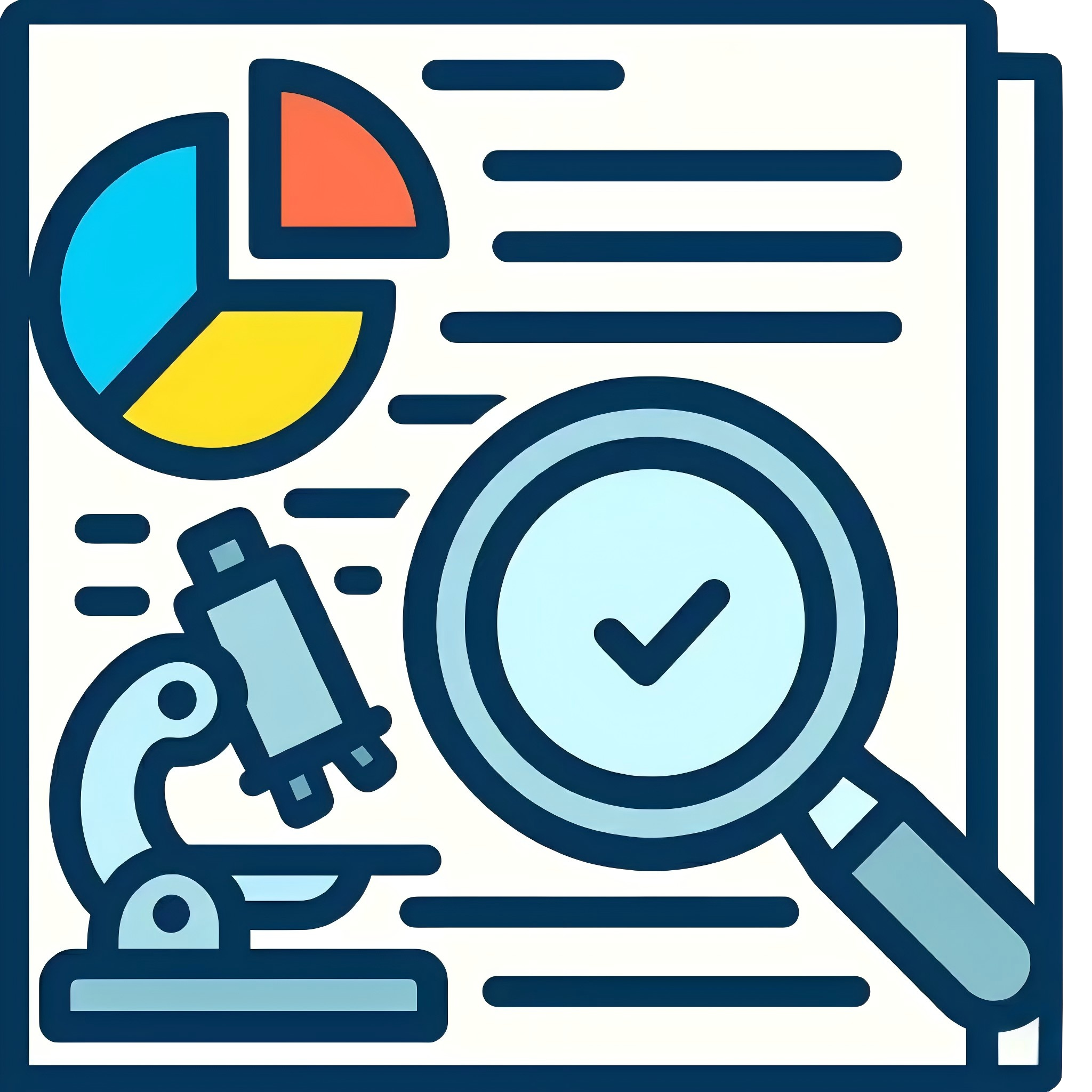}}}
\newcommand{\github}{\raisebox{-1.5pt}{\includegraphics[height=1em]{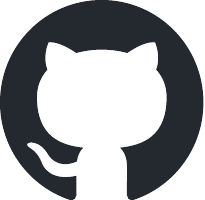}}}
\newcommand{\huggingface}{%
\raisebox{-1.5pt}{\includegraphics[height=1em]{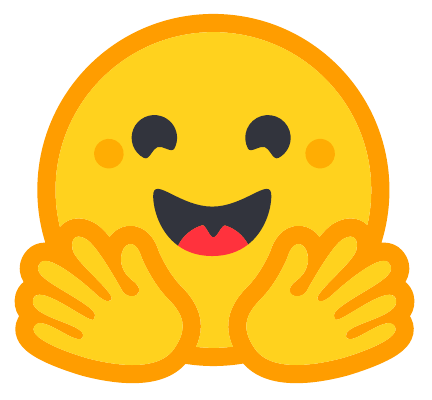}}%
}

\title{\homepage\hspace{0.6em}Decoding \textcolor{SPUR1}{S}cientific Experimental Images:\\
The \textcolor{SPUR1}{S}\textcolor{SPUR2}{P}\textcolor{SPUR3}{U}\textcolor{SPUR4}{R} Benchmark for \textcolor{SPUR2}{P}erception, \textcolor{SPUR3}{U}nderstanding, and \textcolor{SPUR4}{R}easoning}
\author{
    Junpeng Ding \quad
    Zichen Tang \quad
    Haihong E\thanks{Corresponding author.} \quad
    Mengyuan Ji \quad
    Yang Liu \quad
    Haolin Tian \\
    \textbf{Haiyang Sun} \quad
    \textbf{Pengqi Sun} \quad
    \textbf{Yang Xu} \quad
    \textbf{Yichen Liu} \quad
    \textbf{Haocheng Gao} \quad
    \textbf{Zijie Xi} \\
    \textbf{Ruomeng Jiang} \quad
    \textbf{Peizhi Zhao} \quad
    \textbf{Rongjin Li} \quad
    \textbf{Yuanze Li} \quad
    \textbf{Jiacheng Liu} \\
    \textbf{Zhongjun Yang} \quad
    \textbf{Jintong Chen} \quad
    \textbf{Siying Lin}\\
Beijing University of Posts and Telecommunications\\
\homepage~~\href{https://bupt-reasoning-lab.github.io/SPUR}{\texttt{bupt-reasoning-lab.github.io/SPUR}}\\
\github~~\href{https://github.com/BUPT-Reasoning-Lab/SPUR}{\texttt{BUPT-Reasoning-Lab/SPUR}} 
\quad
\huggingface~~\href{https://huggingface.co/datasets/BUPT-Reasoning-Lab/SPUR}{\texttt{BUPT-Reasoning-Lab/SPUR}}
}

\begin{document}
\maketitle



\begin{abstract} 

We introduce \textbf{\ours}, a comprehensive benchmark for scientific experimental image \emph{perception}, \emph{understanding}, and \emph{reasoning}, comprising 4,264 question-answering (QA) pairs derived from 1,084 expert-curated images. \ours features three key innovations: (1) \textbf{Panel-Level Fine-Grained Perception}: evaluating the visual perception of multimodal large language models (MLLMs) across three dimensions (numerical, morphological, and information localization) on six fine-grained panel types; (2) \textbf{Cross-Panel Relation Understanding}: utilizing complex images with an average of 14.3 panels per sample to evaluate MLLMs' ability to decipher intricate cross-panel relations; (3) \textbf{Expert-Level Reasoning}: assessment of qualitative and quantitative reasoning across five experimental paradigms to determine if models can infer conclusions from evidence as human experts do. Comprehensive evaluation of 20 MLLMs and four multimodal Chain-of-Thought (MCoT) methods reveals that current models fall significantly short of the expert-level requirements for scientific image interpretation, underscoring a critical bottleneck in AI for Science (AI4S) research.

\end{abstract}

\section{Introduction}

\begin{figure}[!t]\centering
	\includegraphics[width=0.45\textwidth]{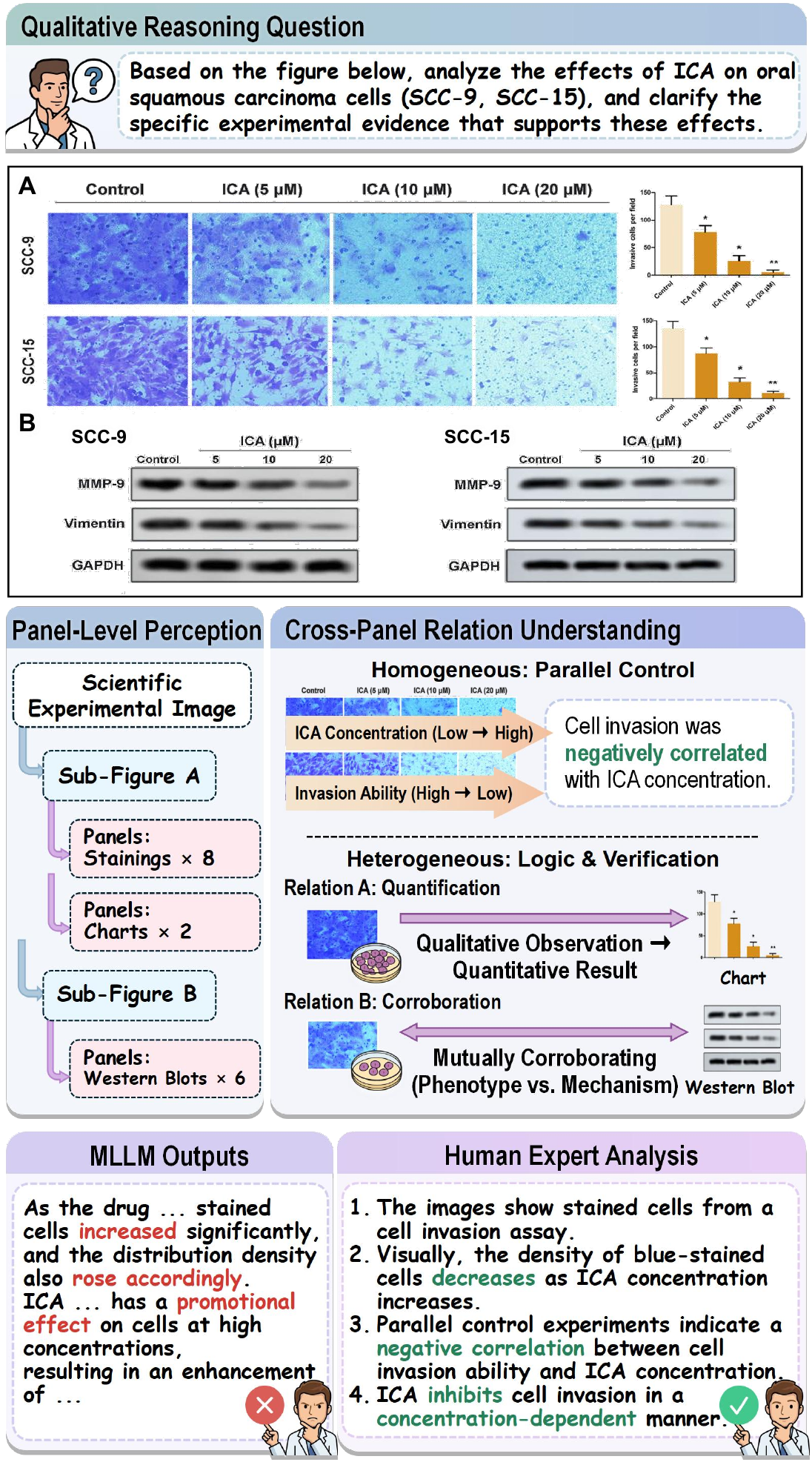}

    \caption{Qualitative Reasoning (Qual.) on a complex multi-panel image. To reach correct conclusions, models must perform (1) panel-level fine-grained perception of morphological features, (2) cross-panel relation understanding, and (3) expert-level reasoning. Current MLLMs struggle with visual trends, leading to hallucinated conclusions compared to human expert analysis.}

    \label{fig: Example}
\end{figure}

\begin{table*}[htbp]
\centering
\resizebox{1.0\textwidth}{!}{%
\renewcommand{\arraystretch}{1.1}

\begin{tabular}{l c c c c c c c c c c }
\toprule
\multirow{2}{*}{\textbf{Benchmark}} &
\multirow{2}{*}{\textbf{Domain}} &
\multirow{2}{*}{\textbf{\# Samples (k)}} &
\multicolumn{2}{c}{\textbf{Panel-Level Perception}} & 
\multicolumn{3}{c}{\textbf{Cross-Panel Understanding}} &
\multicolumn{3}{c}{\textbf{Expert-Level Reasoning}} \\
\cmidrule(lr){4-5} \cmidrule(lr){6-8} \cmidrule(lr){9-11}
& & & Gran. & Dimension & \# Panels/Image & Desc. & Rel.  & Paradigm & Qual. & Quant. \\
\midrule
\multicolumn{11}{l}{\textit{Non-Academic Images}} \\
ScienceQA   & Science & 21.2 & Figure  & N+IL  & 1.0 & \textcolor{my_red}{\ding{56}}  & \textcolor{my_red}{\ding{56}}  & VQA & \textcolor{my_green}{\ding{52}} & \textcolor{my_red}{\ding{56}} \\
MMMU        & Science & 11.5 & Figure  & N+IL  & 2.5 & \textcolor{my_red}{\ding{56}}  & \textcolor{my_red}{\ding{56}}  & VQA & \textcolor{my_green}{\ding{52}} & \textcolor{my_red}{\ding{56}} \\
M3CoT       & Science & 11.3 & Figure  & N+IL  & 1.0 & \textcolor{my_red}{\ding{56}} & \textcolor{my_red}{\ding{56}} & VQA & \textcolor{my_green}{\ding{52}} & \textcolor{my_red}{\ding{56}} \\
EMMA        & Science & 2.7 & Figure  & N+IL  & 1.3 & \textcolor{my_red}{\ding{56}} & \textcolor{my_red}{\ding{56}}  & VQA & \textcolor{my_green}{\ding{52}} & \textcolor{my_green}{\ding{52}} \\
\bottomrule
\multicolumn{11}{l}{\textit{Academic Images}} \\
MMSci       & Science & 742.3 & Sub-fig.  & N+IL & 7.4 & \textcolor{my_green}{\ding{52}} & \textcolor{my_red}{\ding{56}}  & Caption & \textcolor{my_red}{\ding{56}} & \textcolor{my_red}{\ding{56}} \\
SciAssess   & Science & 6.9 & Figure  & M+IL & 3.4 & \textcolor{my_red}{\ding{56}}& \textcolor{my_red}{\ding{56}}  & VQA & \textcolor{my_green}{\ding{52}} & \textcolor{my_red}{\ding{56}} \\
SFE         & Science & 0.8 & Sub-fig.  & N+M+IL  & 2.3 & \textcolor{my_green}{\ding{52}}& \textcolor{my_red}{\ding{56}} & VQA & \textcolor{my_green}{\ding{52}}& \textcolor{my_red}{\ding{56}} \\

Text2Analysis  & Statistics & 2.2 & Figure  & N+IL & 1.0 & \textcolor{my_red}{\ding{56}}& \textcolor{my_red}{\ding{56}}  & VQA & \textcolor{my_green}{\ding{52}} & \textcolor{my_green}{\ding{52}} \\
EvoChart       & Statistics & 1.2 & Figure  &  N+IL  & 1.0 & \textcolor{my_red}{\ding{56}}& \textcolor{my_red}{\ding{56}}  & VQA & \textcolor{my_green}{\ding{52}} & \textcolor{my_green}{\ding{52}} \\
MISS-QA        & CS & 1.5 & Sub-fig.  &  N+IL  & 1.6 & \textcolor{my_red}{\ding{56}}& \textcolor{my_red}{\ding{56}}  & VQA & \textcolor{my_green}{\ding{52}} & \textcolor{my_red}{\ding{56}} \\
SPIQA          & CS & 270 & Sub-fig.  &  N+IL  & 2.7 & \textcolor{my_red}{\ding{56}}& \textcolor{my_red}{\ding{56}}  & VQA   & \textcolor{my_green}{\ding{52}} & \textcolor{my_red}{\ding{56}} \\
OmniMedVQA     & Medicine & 127.9 & Sub-fig.  & M+IL   & 1.0 & \textcolor{my_red}{\ding{56}}& \textcolor{my_red}{\ding{56}} & VQA & \textcolor{my_green}{\ding{52}} & \textcolor{my_red}{\ding{56}} \\
MicroVQA       & Medicine & 1.0 & Panel  & N+M  & 1.9 & \textcolor{my_green}{\ding{52}} & \textcolor{my_green}{\ding{52}}  & Exp. Design & \textcolor{my_green}{\ding{52}} & \textcolor{my_red}{\ding{56}} \\
\midrule
\textbf{\ours (ours)}  & \textbf{Experiment} & \textbf{4.3} & \textbf{Panel} & \textbf{N+M+IL} & \textbf{14.3} & \textcolor{my_green}{\ding{52}} & \textcolor{my_green}{\ding{52}}  & \textbf{Exp. Reason} & \textbf{36.1\%} & \textbf{63.9\%} \\

\bottomrule
\end{tabular}

}

\caption{Comparison of \ours and related benchmarks. For \emph{perception} features, \textbf{Gran.}: Granularity; \textbf{N}: Numerical; \textbf{M}: Morphological; \textbf{IL}: Information Localization. For \emph{understanding} capabilities, \textbf{Desc.}: Information Description; \textbf{Rel.}: Relation Understanding. For \emph{reasoning} tasks, \textbf{Qual.}: Qualitative; \textbf{Quant.}: Quantitative.}  
\label{tab:dataset_comparison}
\end{table*}

Recently, multimodal large language models (MLLMs) have demonstrated remarkable capabilities in \textbf{\emph{perceiving}} and \textbf{\emph{understanding}} scientific images across diverse domains, including statistical charts~\cite{MultimodalArXiv,SciFiBench,Charxiv}, tables~\cite{Multimodal}, biomedical images~\cite{Micro-bench}, model schematics~\cite{Microvqa}, and chemical diagrams~\cite{Chemvlm}. This establishes their profound potential for AI for Science (AI4S)~\cite{DeepResearchSurvey}. The multimodal \textbf{\emph{reasoning}} capacities of MLLMs are further enhanced through multimodal Chain-of-Thought (MCoT) methods, which amplify structured reasoning via (1) prompt-based methods~\cite{DDCoT,VisOfThought}, (2) plan-based decomposition~\cite{zheng2024thinkinglookingimprovingmultimodal, CantorMultimodalCoT}, and (3) training-based frameworks~\cite{OrderChain, VLMCoTReasoning}.


Despite these advancements, complex multi-panel scientific experimental images, which are ubiquitous in scholarly publications, remain underexplored. As shown in Figure~\ref{fig: Example}, such images systematically arrange multiple panels to present experimental processes and results. MLLMs are expected to emulate human experts in achieving: (1) \textbf{fine-grained perception} of individual panels, (2) \textbf{cross-panel understanding} of intricate relations, and (3) \textbf{qualitative
and quantitative reasoning} toward scientific conclusions. However, as summarized in Table~\ref{tab:dataset_comparison}, current scientific reasoning benchmarks exhibit critical limitations in meeting real-world AI4S demands:


\begin{itemize} [leftmargin=*]
\item \textbf{Absence of Scientific Experimental Images.}
Experimental images (\eg{ cellular staining images and Western blots}) exhibit unique visual characteristics that pose severe fine-grained perception challenges. While benchmarks like ScienceQA~\cite{NEURIPS2022_11332b6b}, M3CoT~\cite{M3CoT-M3}, and MMMU~\cite{Mmmu} aggregate diverse scientific imagery for VQA tasks, they severely underrepresent experimental images.

\item \textbf{Lack of Cross-Panel Relation Understanding.}
Interpreting experimental images requires comprehending panel correlations (\eg{ control vs. treatment groups}), which is a core capability in scientific workflows. However, existing benchmarks like MMSci~\cite{li2024mmsci}, SciAssess~\cite{SciAssess}, and SFE~\cite{zhou2025scientists} predominantly feature isolated images with limited complexity, averaging fewer than eight panels per sample. 

\item \textbf{Neglect of Quantitative Reasoning.} Quantitative reasoning, involving precise analysis of proportions, magnitudes, and metrics, poses substantially greater challenges than qualitative conclusion judgments. Although pioneering works address qualitative aspects (\eg{ experimental design in MicroVQA~\cite{Microvqa}}), they largely neglect MLLMs' quantitative reasoning capabilities for deriving scientific conclusions.

\end{itemize}

\begin{figure*}[t]
	\centering
	\includegraphics[width=0.98\textwidth]{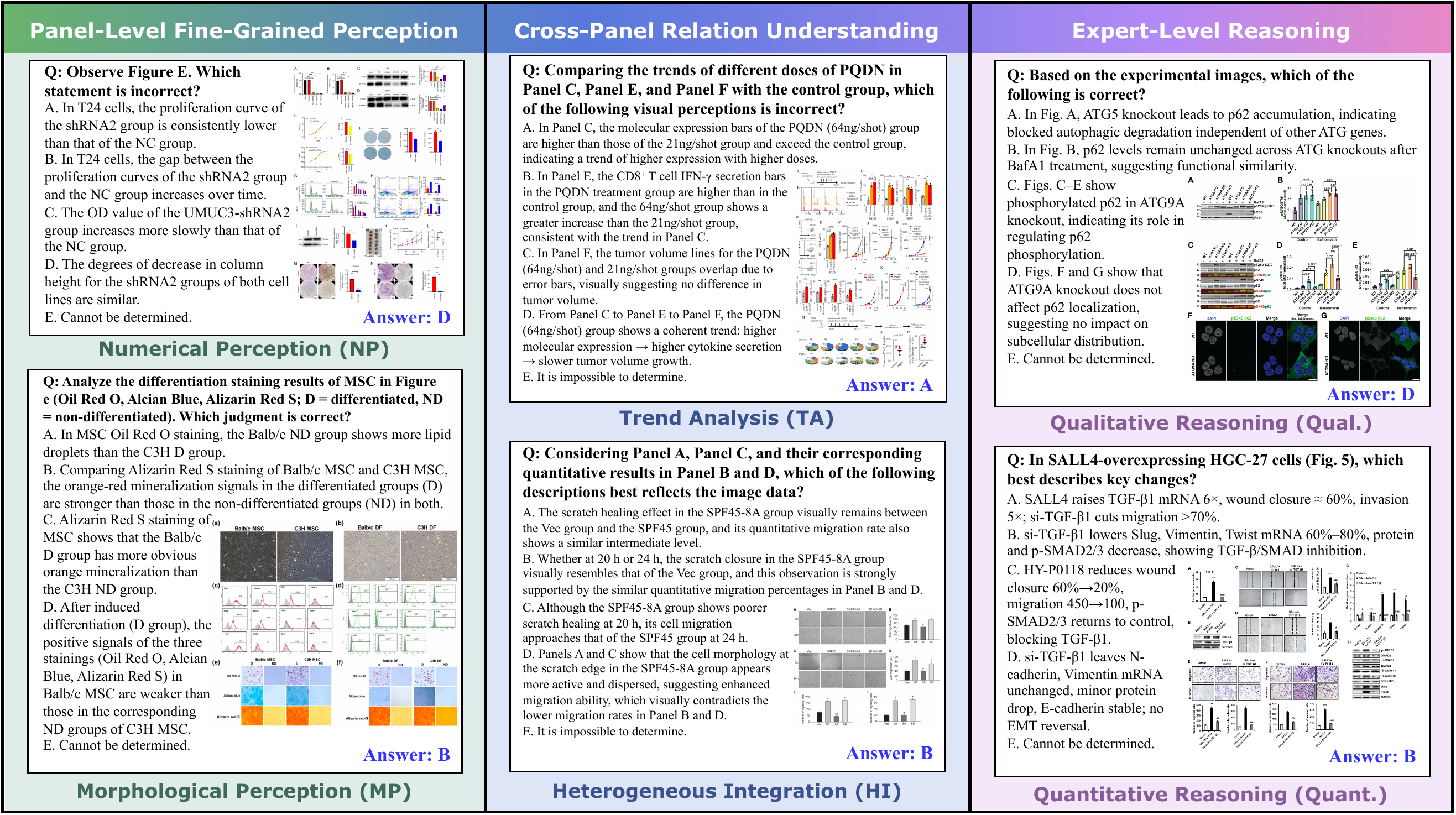} 
	\caption{Representative QA pairs across \ours's three cognitive stages. See Appendix~\ref{Appdix.a} for more examples.}

    \label{fig:benchmark_overview}
\end{figure*}

To bridge these gaps, we propose \textbf{\ours}, a benchmark for multimodal \textbf{\emph{perception}}, \textbf{\emph{understanding}}, and \textbf{\emph{reasoning}} on scientific experimental images (Figure~\ref{fig:benchmark_overview}), comprising 1,084 images and 4,264 QA pairs. \ours offers three key advantages:

\begin{itemize} [leftmargin=*]
\item \textbf{Cross-Disciplinary Generalization.} 
\ours is the first MLLM benchmark dedicated to scientific experimental image reasoning, filling a critical gap in AI4S evaluation. Distinct from existing benchmarks, it exclusively sources high-quality experimental images from open-access PubMed Central (PMC)\footnote{\url{https://pmc.ncbi.nlm.nih.gov}} papers, covering seven diverse disciplines and five classic experimental paradigms. Notably, these images share consistent visual structures, layout rules, and reasoning logic across fields, ensuring strong cross-disciplinary commonality.

\item \textbf{Fine-Grained Panels and Complex Cross-Panel Relations.} 
\ours utilizes highly complex images containing an average of 14.3 panels per sample. These are categorized into six fine-grained types (\ie{ charts, Western blots, and four subtypes of staining images}). Further, we model cross-panel relations, including isomorphic relations (\eg{ dose-response trends across staining images}) and heterogeneous relations (\eg{ Western blot vs. subcellular staining co-validation}).

\item \textbf{Multi-Dimensional Evaluation Framework.} 
We establish a three-stage evaluation framework (\emph{Perception} $\rightarrow$ \emph{Understanding} $\rightarrow$ \emph{Reasoning}) with seven core tasks: 
(1) The \emph{Perception} stage focuses on \textbf{panel-level analysis}, evaluating Numerical Perception (NP, \eg{ estimate kinetic curve values}), Morphological Perception (MP, \eg{ cell invasion extent identification}), and Information Localization (IL); 
(2) The \emph{Understanding} stage deciphers \textbf{cross-panel relations}, testing Trend Analysis (TA) for isomorphic panels and Heterogeneous Integration (HI) across diverse panels; 
(3) The \emph{Reasoning} stage requires \textbf{expert-level judgments}, assessing both Qualitative Reasoning (Qual., \eg{ directional conclusions}) and Quantitative Reasoning (Quant.).

\end{itemize}

We evaluate eight proprietary and 12 open-source MLLMs, along with four training-free MCoT methods (\ie{ DDCoT~\cite{DDCoT}, VoT~\cite{VisOfThought}, VIC~\cite{zheng2024thinkinglookingimprovingmultimodal}, and Cantor~\cite{CantorMultimodalCoT}}), revealing three key findings:

\begin{itemize} [leftmargin=*]
\item \textbf{MLLMs underperform in the \emph{perception} $\rightarrow$ \emph{understanding} $\rightarrow$ \emph{reasoning} pipeline.}
Only \texttt{Gemini 3 Pro Preview} exceeds 60\% accuracy, indicating that current models fall significantly short of rigorous AI4S requirements

\item \textbf{Critical deficiencies exist across all evaluation stages.}
Numerical Perception (NP) yields the poorest results (most models $<$ 60\%). Although Morphological Perception (MP) performs better, it exhibits category bias and limited generalization. Cross-Panel Relation Understanding remains a core bottleneck, with accuracy plummeting as relation complexity increases in Trend Analysis (TA). Furthermore, Quantitative Reasoning (Quant.) significantly trails Qualitative Reasoning (Qual.) by 12.76\%--31.41\%, failing to meet expert-level demands.

 \begin{figure*}[t]
	\centering
	\includegraphics[width=0.98\textwidth]{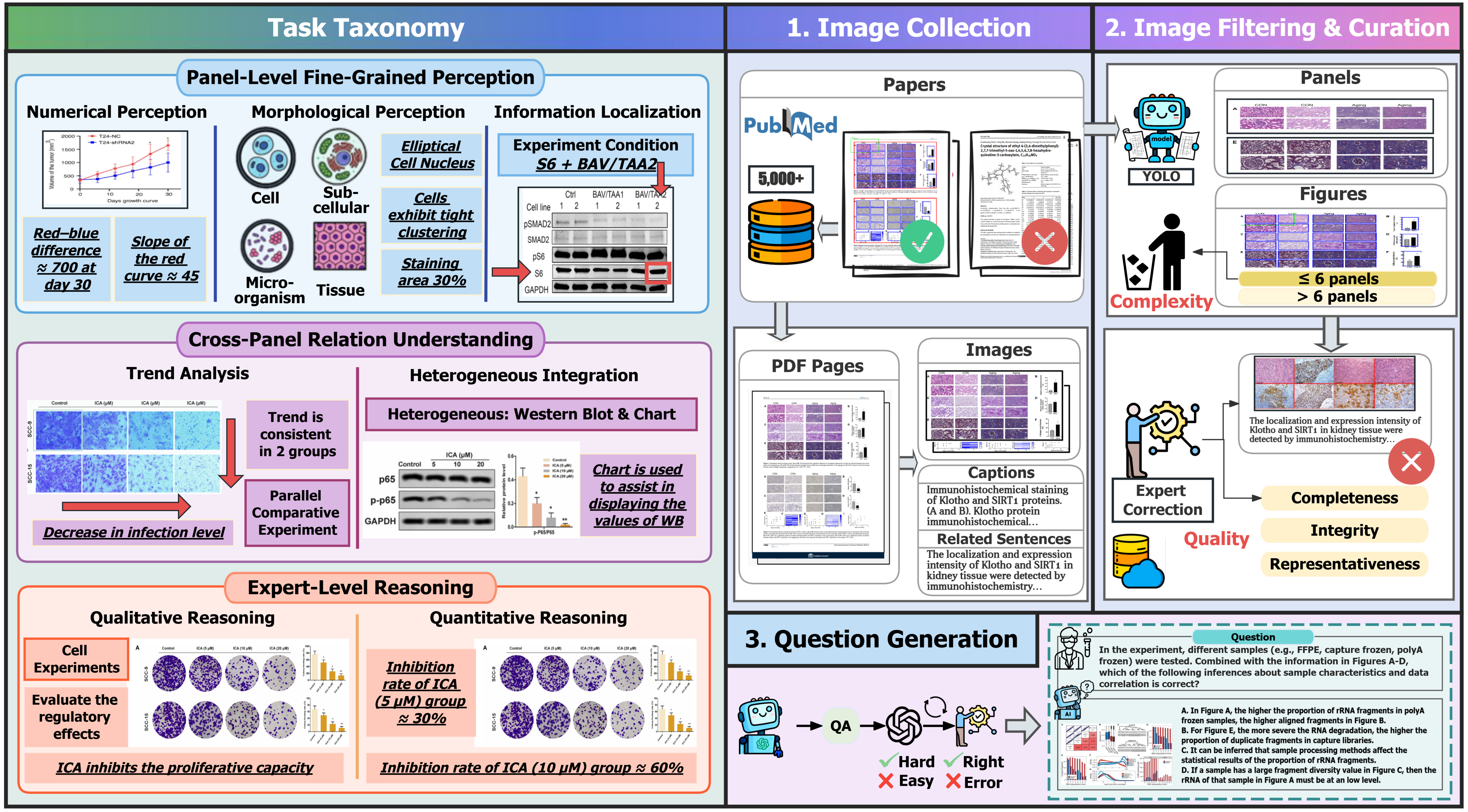}
	\caption{Overview of \ours. \textbf{(Left)} The hierarchical task taxonomy comprising seven core tasks. \textbf{(Right)} The three-stage QA pair curation pipeline.}

	\label{fig:curation_pipeline}
\end{figure*}

\item \textbf{MCoT methods lack consistent gains in AI4S visual reasoning.} 
Prompt-based MCoT methods merely augment reasoning steps without mitigating perceptual limitations. Consequently, they often amplify perceptual hallucinations rather than enhance reasoning outcomes. This indicates that robust panel-level perception is an absolute prerequisite for effective AI4S visual reasoning.

\end{itemize}


\section{SPUR Benchmark}

We introduce \ours to evaluate MLLMs' \textbf{\emph{perception}}, \textbf{\emph{understanding}}, and \textbf{\emph{reasoning}} capabilities on scientific experimental images through seven core tasks. \ours comprises 4,264 multiple-choice questions (MCQs) paired with 1,084 scientific images and texts from PMC, spanning seven disciplines, five experimental paradigms, and six fine-grained panel types.

\subsection{Task Taxonomy}

As shown in Figure~\ref{fig:curation_pipeline}, \ours structures its seven tasks across the \emph{Perception} $\rightarrow$ \emph{Understanding} $\rightarrow$ \emph{Reasoning} stages:

\paragraph{Panel-Level Fine-Grained Perception.}
This stage identifies and parses visual features within individual panels (\eg{ stained preparations}) to establish foundational visual cognition:

\begin{itemize} [leftmargin=*]
\item \textbf{Numerical Perception (NP).} Quantifies absolute levels and differentiation of visual features.
\item \textbf{Morphological Perception (MP).} Analyzes biological structure morphology in stained preparations (\eg{ cell shape and tissue architecture}).
\item \textbf{Information Localization (IL).} Maps panels to their corresponding experimental conditions.
\end{itemize}

\paragraph{Cross-Panel Relation Understanding.} 
This stage extracts implicit relations across multi-panel frameworks (\eg{ causal, comparative, and argumentative patterns}):

\begin{itemize} [leftmargin=*]
\item \textbf{Trend Analysis (TA).} Interprets directional changes and experimental content across isomorphic panels.
\item \textbf{Heterogeneous Integration (HI).} Aligns and reasons across disparate panel types through cross-modal synthesis (\eg{ information integration and abstract-concrete mapping}).
\end{itemize}

\paragraph{Expert-Level Reasoning.} 
This stage derives scientific inferences from multi-panel visual evidence integrated with experimental context:

\begin{itemize} [leftmargin=*]
\item \textbf{Qualitative Reasoning (Qual.).} Synthesizes visual cues, domain knowledge, and experimental design logic to interpret biological significance.
\item \textbf{Quantitative Reasoning (Quant.).} Conducts mathematical verification and evaluation of inter-group differences, effect sizes, and statistical significance via numerical comparison, ratio calculation, and hypothesis testing.
\end{itemize}

\subsection{Dataset Construction and Annotation}
As shown in Figure~\ref{fig:curation_pipeline}, \ours was built through a three-stage process (see Appendix~\ref{Appdix.b} for details):
\paragraph{Image Collection.} 
We systematically curated over 5,000 open-access papers from PMC, applying two selection criteria: publication within the last decade and a source journal impact factor (IF) $>$ 3.0. Using automated PDF parsing, we extracted 5,632 scientific images. These were manually annotated with related sentences, contextual captions, and disciplinary classifications to form the raw dataset.

\paragraph{Image Filtering and Curation.} 
To ensure question complexity, images underwent a dual-filtering process. First, a YOLO-based panel detector segmented images and excluded those with $\leq$ 6 panels (removing 77.6\% of candidates). Second, experts verified experimental workflow completeness and annotation accuracy, discarding 14.2\% of the remaining set for lacking methodological rigor.

\paragraph{Question Generation.} 
Domain experts developed specialized prompt templates aligned with our task hierarchy, enabling \texttt{GPT-4o} to generate 7,608 candidate QA pairs. 

\begin{figure}[t]
	\centering
	\includegraphics[width=0.98\columnwidth]{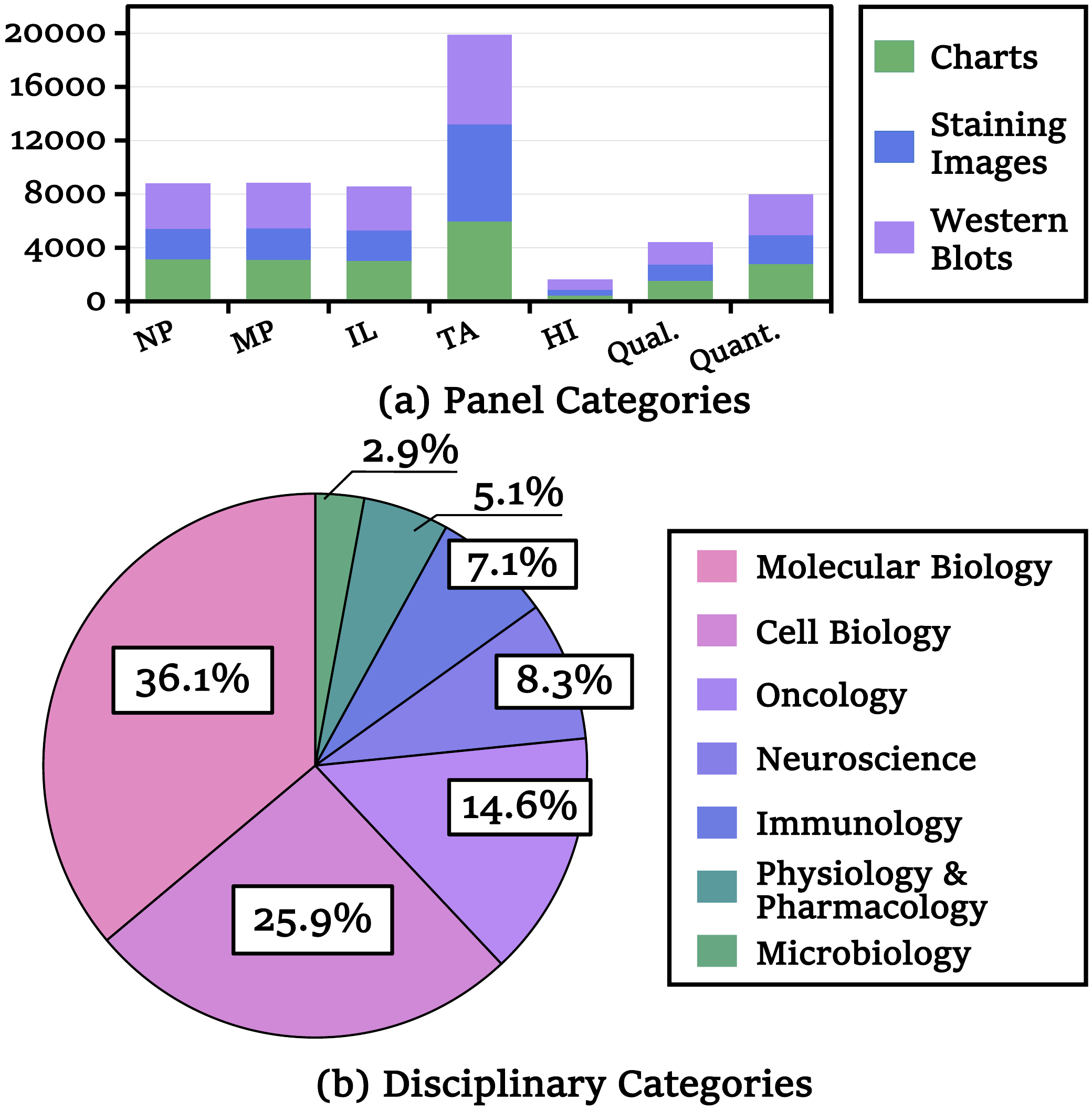}
	\caption{Distribution of \textbf{(a)} Panel Categories and \textbf{(b)} Disciplinary Categories in \ours.}
	\label{fig:distribution}
\end{figure}

\subsection{Data Quality Assurance}

\paragraph{Textual Shortcut Elimination.} 
We implemented a statistical filter to eliminate text-dependent questions. For each QA pair, \texttt{GPT-4o} answered text-only queries 10 times under randomized conditions. Questions with $\geq$ 5 correct responses were discarded (removing 21.2\% of candidates), ensuring the retained items require genuine visual reasoning. Subsequently, the remaining \texttt{(question, options, answer)} triplets underwent expert review for biomedical validity and task alignment, with 28.9\% rejected primarily due to factual inaccuracies or task mismatch.

\paragraph{Expert Review.}
We implemented a rigorous quality control system through standardized expert review protocols. Four domain specialists ($>$ 40 peer-reviewed publications each) and two senior experts ($>$ 100 publications) conducted multi-tiered validation. Each sample underwent independent dual-expert review to flag inadequate information or ambiguous answers, with senior arbitrators resolving any discrepancies.

\paragraph{Mitigating Data Leakage.}
During Question Generation, we strictly prohibited deriving questions and options directly from captions or related text. Instead, they were designed exclusively based on panel visuals, demanding fine-grained perception and cross-panel reasoning. Furthermore, filtering out text-only answerable questions prevents models from relying on pre-trained knowledge, common sense, or superficial textual patterns.

\begin{table}[b!]
\centering
\resizebox{\columnwidth}{!}{
\begin{tabular}{lr}
\toprule
\textbf{Property}                            & \textbf{Value}  \\
\midrule
\multicolumn{2}{c}{\textbf{Data Source}} \\
\noalign{\vskip 1ex}
\# Experimental Paradigms/Images & 5/1,084 \\
Avg. Panels/Image & 14.3 \\
\midrule

\multicolumn{2}{c}{\textbf{Multiple-Choice Question}} \\
\noalign{\vskip 1ex}
\textbf{Total Samples} & \textbf{4,264}\\
\noalign{\vskip 0.5ex}\hdashline\noalign{\vskip 0.5ex}
\quad \# Perception (NP/MP/IL) & 636/634/621\\
\quad \# Understanding (TA/HI) & 1,357/130\\
\quad \# Reasoning (Qual./Quant.) & 567/319\\
\noalign{\vskip 0.5ex}\hdashline\noalign{\vskip 0.5ex}
Avg. Words (Question/Options/Evidence)  & 20.8/54.6/58.8 \\

\bottomrule

\end{tabular}
}
\caption{Key statistics of \ours.}
\label{tab:statistics}
\vspace{-0.2cm}
\end{table}

\begin{table*}[t]
\tablestyle{5pt}{1.1}
\centering
\small
\resizebox{\textwidth}{!}{

\begin{tabular}{l cccc ccc ccc c}
\toprule
\multirow{3}{*}{\textbf{Model}} 
& \multicolumn{4}{c}{\textbf{Panel-Level Perception}} 
& \multicolumn{3}{c}{\textbf{Cross-Panel  Understanding}} 
& \multicolumn{3}{c}{\textbf{Expert-Level Reasoning}} 
&\multirow{3}{*}{\textbf{Overall}}\\
\cmidrule(lr){2-5} 
\cmidrule(lr){6-8}
\cmidrule(lr){9-11}
& \textbf{NP}& \textbf{MP}& \textbf{IL}& \textbf{Micro Avg.}& \textbf{TA}& \textbf{HI}& \textbf{Micro Avg.}& \textbf{Qual.}& \textbf{Quant.}& \textbf{Micro Avg.}\\

\midrule
\noalign{\vspace{-0.5ex}}
\rowcolor{gray!8}\multicolumn{12}{c}
{\textbf{\textit{Proprietary MLLMs}}}\\
\midrule
Gemini 3 Pro Preview  & \best{\textbf{61.26}} & \best{\textbf{67.74}} & \best{\textbf{59.67}} & \best{\textbf{62.92}} & 51.04 & 59.23 & 51.77 & \best{\textbf{90.31}} & \second{58.90}& \best{\textbf{70.29}} &\best{\textbf{60.57}}\\
Claude 3.7 Sonnet (thinking)   & \second{59.67} & 64.32 & 57.45 & \second{60.50} & 51.30 & 60.80 & 52.12 & \second{87.58} & \best{\textbf{59.96}}& \second{69.93}  & 59.52\\
Gemini 2.5 Pro Preview   & 56.47 & 62.97 & 56.47 & 58.65 & 53.30 & 61.54 & 54.02 & 86.54 &  57.94 & 68.24 & 59.00\\
GPT-5.1   & 58.73 & 61.72 & 54.47 & 58.33 & 51.18 & 50.78 & 51.15 & 86.52 & 56.36 & 67.23 &57.68\\
OpenAI o4-mini-high  & 59.24 & \second{64.50} & 55.34 & 59.72 & 48.37 & 59.23 & 49.32 & 84.33 & 56.36 & 66.44 &57.50\\
Doubao-Seed-1.6   & 53.30 & 63.51 &  56.61 & 57.81 & 47.83 & 56.92 & 48.62 & 80.31 & 53.89 & 63.43 &55.77\\
GPT-4o   & 45.91 & 53.81 & 52.27 & 50.64 & 53.21 & 63.57 & 54.12 & 71.16 & 54.80 & 60.73 &53.95\\
Grok 4.1 Fast  & 47.33 & 55.99 & 52.98 & 52.09 & 43.70 & 52.31 & 44.45 & 73.44 & 48.94 & 57.79 & 50.61\\

\midrule
\noalign{\vspace{-0.5ex}}
\rowcolor{gray!8}\multicolumn{12}{c}
{\textbf{\textit{Open-Source MLLMs}}}\\
\midrule
GLM-4.5V  & 57.70 & 61.99 & \second{57.65} & 59.12 & 55.71 & \second{68.46} & 56.83 & 80.94 &  58.48 & 66.59 & \second{59.87}\\
Ministral 3 14B  & 50.88 & 61.40 & 56.56 & 56.25 & \best{\textbf{57.79}} & \best{\textbf{70.00}} & \best{\textbf{58.86}} & 72.50 & 56.81 & 62.49 & 58.48\\

Ministral 3 8B  & 51.57 & 57.03 & 57.03 & 55.19 & \second{57.49} & 66.15 & \second{58.25} & 70.85 &53.53 &59.77 & 57.21\\
Llama 4 Maverick  & 51.73 & 59.78 & 56.61 & 56.03 & 51.88 & 58.46 & 52.46 & 84.64 & 57.02 & 67.01 & 57.06\\
Qwen3-VL-30B-A3B-Thinking & 53.48 & 58.00 & 55.27 & 53.99 & 54.85 & 61.24 & 55.41 & 75.16 & 54.17 & 61.75 & 56.10\\
InternVL3-78B & 46.30 & 51.97 & 49.84 & 49.36 & 49.52 & 61.24 & 50.54 & 75.24 & 51.06 & 59.80 & 51.94\\
Mistral Small 3.1  & 42.74 & 51.92 & 53.48 & 49.36 & 51.82 & 56.92 & 52.28 & 72.82 & 47.56 & 56.84 & 51.94\\
Qwen3-VL-30B-A3B-Instruct & 44.18 & 53.31 & 51.05 & 49.50 & 50.41 & 63.08 & 51.51 & 66.88 & 51.41 & 57.00 & 51.76\\
Gemma 3 27B & 38.36 & 46.43 & 48.30 & 44.33 & 48.59 & 57.69 & 49.39 & 64.14 & 51.38 &  55.95  &48.44\\
Qwen2.5-VL-72B & 38.10 & 45.34  & 49.11 & 44.38 & 51.87 & 61.90 & 52.82 & 73.10 & 52.51 & 59.95  &48.21\\
LLaVA-v1.5-13B & 33.05 & 28.11 & 34.15 & 31.75 & 34.52 & 44.96 & 35.43 & 62.19 & 35.58  & 45.20  &35.97\\
LLaVA-OneVision-7B & 26.22 & 35.71 & 40.68 & 31.49 & 34.14 & 40.63 & 34.79 & 48.59 & 35.65 & 40.34 &34.47\\

\bottomrule
\end{tabular}
}
\caption{Performance comparison of proprietary and open-source MLLMs across three cognitive stages and seven core tasks on \ours. The best and second-best scores are highlighted in \colorbox{ForestGreen!30}{\textbf{bold}} and \colorbox{blue!15}{\underline{underlined}}, respectively.}
\label{tab:main_result}
\end{table*}


\subsection{Data Statistics}

As shown in Table~\ref{tab:statistics} and Figure~\ref{fig:distribution}, \ours comprises 4,264 expert-curated MCQs derived from 1,084 images, encompassing over 15,500 panels across seven disciplines. Averaging 14.3 panels per image across six fine-grained types, it enables complex cross-panel relations that significantly surpass existing benchmarks in structural complexity. Task distribution includes 636 NP, 634 MP, 621 IL, 1,357 TA, 130 HI, 567 Qual., and 319 Quant. samples. The benchmark's high difficulty is reflected by lengthy options (averaging 54.6 words) and detailed reasoning evidence (58.8 words), demanding robust multimodal capabilities well beyond general benchmarks (See Appendix~\ref{appendix.diversity} for more details).


\section{Experiments}

\subsection{Experimental Setup}
\paragraph{Models.} 
We evaluate 20 state-of-the-art MLLMs with visual capabilities (eight proprietary and 12 open-source), spanning diverse architectural paradigms. Model details are provided in Appendix~\ref{Appdix.c.1}.

\paragraph{Metrics.} 
Model performance is assessed by accuracy on MCQs across all seven tasks.

\paragraph{MCoT Methods.}
We evaluate two categories of MCoT methods encompassing four specific approaches: 
(1) \textit{Prompt-based} methods (\ie{ DDCoT~\cite{DDCoT} and VoT~\cite{VisOfThought}}) using designed prompts for rationale generation; 
(2) \textit{Plan-based} methods (\ie{ VIC~\cite{zheng2024thinkinglookingimprovingmultimodal} and Cantor~\cite{CantorMultimodalCoT}}) enabling dynamic thought exploration during inference.

\paragraph{Research Questions.} 
We investigate five core questions: 
\textbf{RQ1}: Do MLLMs exhibit expert-level perception-to-reasoning capabilities on scientific experimental images? 
\textbf{RQ2}: Can MLLMs achieve precise panel-level perception? 
\textbf{RQ3}: How effectively do MLLMs understand complex cross-panel relations? 
\textbf{RQ4}: Do MLLMs perform expert-level qualitative and quantitative reasoning? 
\textbf{RQ5}: Do MCoT methods enhance scientific reasoning performance?

\subsection{Overall Performance (RQ1)}
Table~\ref{tab:main_result} presents the comprehensive results. Our main findings are summarized as follows:

\noindent \textbf{\emph{Overall model performance remains inadequate.}}
Except for \texttt{Gemini 3 Pro Preview}, all models achieve $<$ 60\% accuracy on \ours. This substantial gap indicates that MLLMs' perception, understanding, and reasoning capabilities fall far below expert-level requirements for AI4S tasks.

\noindent \textbf{\emph{Limited panel-level perception constrains reasoning.}}
Open-source models consistently underperform proprietary counterparts in fine-grained perception tasks. All models scoring $<$ 63\% highlights fundamental perceptual limitations across mainstream MLLMs.

\noindent \textbf{\emph{Cross-panel understanding presents critical bottlenecks.}}
Models attain significantly lower accuracy in Trend Analysis (TA) than in other tasks, with most showing the poorest performance here. This underscores that comprehending complex cross-panel relations remains a core challenge.

\noindent \textbf{\emph{Reasoning tasks reveal substantial disparities.}}
While overall reasoning accuracy exceeds understanding tasks, Quantitative Reasoning (Quant.) lags Qualitative Reasoning (Qual.) by 12.76\%--31.41\%. This confirms that current MLLMs cannot meet rigorous quantitative analysis demands.

 \begin{figure}[!t]
	\centering
	\includegraphics[width=0.98\columnwidth]{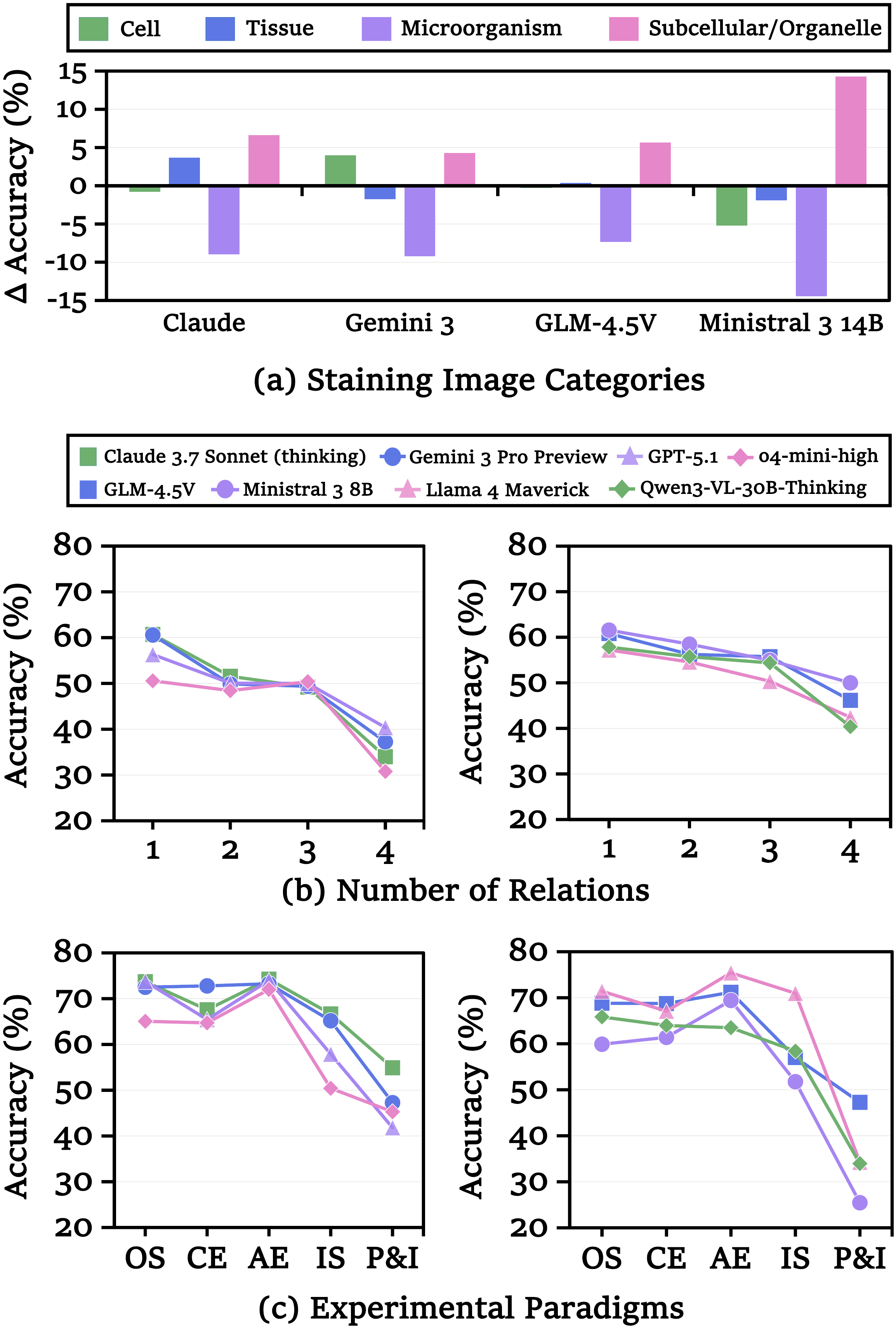}
	\caption{Fine-grained results based on (a) Staining Image Categories, (b) Number of Relations, and (c) Experimental Paradigms. For (a), the Y-axis represents the absolute accuracy deviation from the model's Micro Avg. within the Panel-Level Fine-Grained Perception stage. Further details are provided in Appendix~\ref{Appdix.c.2}.}
    \label{fig:RQ234}
\end{figure}

\subsection{Panel-Level Perception Results (RQ2)}
\noindent \textbf{\emph{Numerical Perception (NP) shows the weakest performance.}}
Most MLLMs struggle with precise numerical content. Among proprietary models, \texttt{Gemini 3 Pro Preview} leads at 61.26\%, while for open-source ones, only \texttt{GLM-4.5V} (57.70\%) approaches mid-tier proprietary performance. This deficiency constrains quantitative reasoning.

\noindent \textbf{\emph{Weak Information Localization (IL) impedes understanding.}}
IL yields the lowest scores among perception tasks, where even the top-performing \texttt{Gemini 3 Pro Preview} achieves only 59.67\%. This prevents accurate panel-to-condition linking, hindering cross-panel relation construction and subsequent reasoning.

\noindent \textbf{\emph{Morphological Perception (MP) exhibits high but unstable accuracy.}}
As shown in Figure~\ref{fig:RQ234}(a), fine-grained analysis across four staining categories reveals a significant imbalance. For instance, \texttt{Ministral 3 14B} scores 70.52\% on Subcellular/Organelle images but only 42.80\% for Microorganisms. Such fluctuations indicate limited generalizability and training data biases.

\subsection{Cross-Panel Understanding Results (RQ3)}
\noindent \textbf{\emph{Models struggle with Trend Analysis (TA) in isomorphic panels.}}
Figure~\ref{fig:RQ234}(b) confirms that TA accuracy inversely correlates with relation complexity. As the number of relations increase from one to four, \texttt{Claude 3.7 Sonnet (thinking)}'s accuracy drops from 60.70\% to 34.00\%, reflecting an inability to handle multi-relation logic.

\noindent \textbf{\emph{Heterogeneous Integration (HI) yields significantly higher performance.}}
This stems from academic contexts: relations across disparate experiments typically involve supportive confirmatory connections, whereas isomorphic panels often generate diverse, contradictory trends, creating inherently more challenging questions.

\begin{figure}[!b]
    \centering
    \includegraphics[width=0.98\columnwidth]{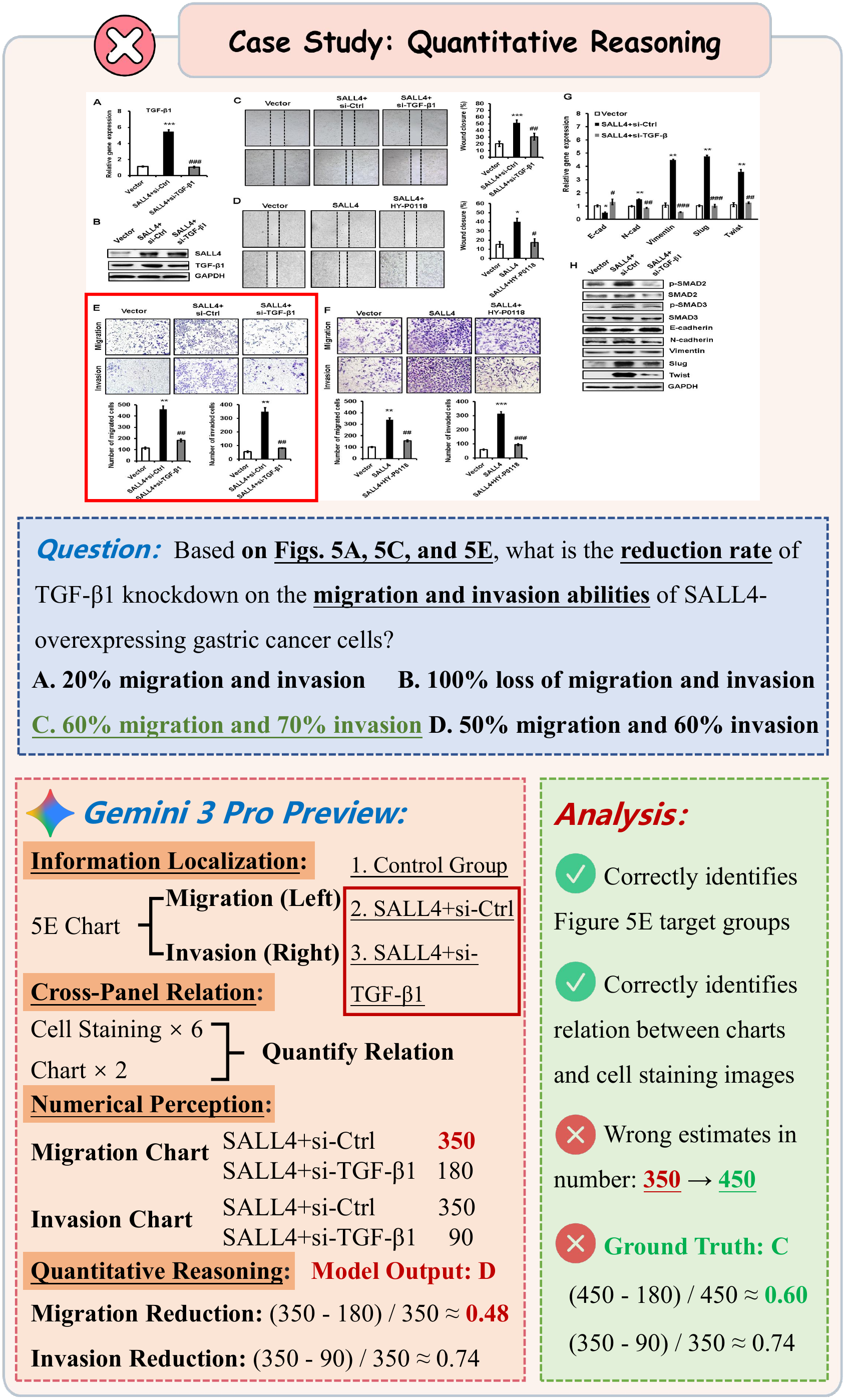}
    \caption{Error analysis of a Quantitative Reasoning (Quant.) task. This case demonstrates how numerical perception errors by \texttt{Gemini 3 Pro Preview} propagate into cascaded reasoning failures.}
    \label{fig: case}
\end{figure}


\begin{table*}[t]
\centering
\scriptsize
\resizebox{\textwidth}{!}{
\begin{tabular}{l cccc ccc ccc c}
\toprule
\multirow{3}{*}{\textbf{Model}} &
\multicolumn{4}{c}{\textbf{Panel-Level Perception}} &
\multicolumn{3}{c}{\textbf{Cross-Panel Understanding}} &
\multicolumn{3}{c}{\textbf{Expert-Level Reasoning}} &
\multirow{3}{*}{\textbf{Overall}} \\
\cmidrule(lr){2-5}\cmidrule(lr){6-8}\cmidrule(lr){9-11}
& \textbf{NP} & \textbf{MP} & \textbf{IL} & \textbf{Micro Avg.} & \textbf{TA} & \textbf{HI} & \textbf{Micro Avg.} & \textbf{Qual.} & \textbf{Quant.} & \textbf{Micro Avg.}\\
\midrule

\noalign{\vspace{-0.5ex}}
\rowcolor{gray!8}\multicolumn{12}{c}
{\textbf{\textit{GLM-4.5V}}}\\
\midrule
Direct
& \textbf{57.70} & 61.99 & \textbf{57.65} & \textbf{59.12} & \textbf{55.71} & \textbf{68.46} & \textbf{56.83} & \textbf{80.94} & \textbf{58.48} & \textbf{66.59} & \textbf{59.87}\\
DDCoT
& 47.11 & 47.67 & 42.83 & 45.88 & 45.24 & 56.35 & 46.19 & 71.52 & 53.27 & 59.93 & 48.90\\
VoT
& 55.82 & \textbf{62.30} & \textbf{57.65} & 58.59 & 53.65 & 60.77 & 54.27 & 78.44 & 57.77 & 65.23 & 58.47\\
VIC
& 35.50 & 33.87 & 32.09 & 33.83 & 27.20 & 29.92 & 27.43 & 34.59 & 36.52 & 35.83 & 32.02\\
Cantor
& 53.41 & 58.69 & 52.96 & 55.04 & 51.23 & 60.00 & 52.00 & 77.12 & 56.61 & 64.05 & 55.59\\
\midrule

\noalign{\vspace{-0.5ex}}
\rowcolor{gray!8}\multicolumn{12}{c}
{\textbf{\textit{Ministral 3 14B}}}\\
\midrule
Direct
& \textbf{50.88} & \textbf{61.40} & \textbf{56.56} & \textbf{56.25} & \textbf{57.79} & \textbf{70.00} & \textbf{58.86} & 72.50 & \textbf{56.81} & \textbf{62.49} & \textbf{58.48} \\
DDCoT
& 47.22 & 52.30 & 47.97 & 49.17 & 51.75 & 61.24 & 52.58 & \textbf{74.61} & 48.48 & 57.97  & 52.19\\
VoT
& 50.00 & 59.05 & 55.68 & 54.89 & 56.67 & 67.97 & 57.66 & 71.92 & 55.14 & 61.24 & 57.17\\
VIC
& 40.10 & 40.00 & 37.34 & 39.16 & 36.93 & 43.08 & 37.47 & 52.50 & 42.55 & 46.15 & 40.02\\
Cantor
& 45.32 & 49.92 & 45.94 & 47.06 & 37.77 & 43.41 & 38.27 & 59.55 & 49.36 & 53.07 & 45.23\\
\midrule

\noalign{\vspace{-0.5ex}}
\rowcolor{gray!8}\multicolumn{12}{c}
{\textbf{\textit{InternVL3-78B}}}\\
\midrule
Direct
& 46.30 & 51.97 & 49.84 & 49.36 & 49.52 & 61.24 & 50.24 & \textbf{75.24} & \textbf{51.06} & \textbf{59.80} & 51.94\\
DDCoT
& 47.86 & 54.62 & \textbf{55.05} & 52.84 & 51.71 & \textbf{62.20} & 52.62 & 70.74 & 50.71 & 57.86 & 53.64\\
VoT
& 47.08 & 52.93 & 51.46 & 50.48 & 50.15 & 62.02 & 51.18 & 73.67 & 50.00 & 58.57 & 52.40\\
VIC
& 40.16 & 46.60 & 40.97 & 42.58 & 39.51 & 48.84 & 40.33 & 55.62 & 39.29 & 45.20 & 42.35\\
Cantor
&\textbf{49.53} & \textbf{56.87} & 54.27 & \textbf{53.54} & \textbf{53.87} & 60.00 & \textbf{54.40} & 71.75 & 47.32 & 56.11 & \textbf{54.37}\\
\midrule

\noalign{\vspace{-0.5ex}}
\rowcolor{gray!8}\multicolumn{12}{c}
{\textbf{\textit{Qwen3-VL-30B-A3B-Instruct}}}\\
\midrule
Direct
& 44.18 & 53.31 & 51.05 & 49.50 & 50.41 & \textbf{63.08} & 51.51 & 66.88 & 51.41 & 57.00 & 51.76\\
DDCoT
& 41.04 & 47.00 & 45.41 & 44.77 & 43.58 & 60.47 & 45.05 & 63.95 & 43.79 & 51.08 & 46.04\\
VoT
& 46.70 & \textbf{56.08} & \textbf{53.95} & \textbf{52.22} & \textbf{51.25} & 60.77 & \textbf{52.09} & \textbf{68.44} & \textbf{51.59} & 57.67 & \textbf{53.31}\\
VIC
& 39.08 & 43.46 & 40.77 & 41.12 & 38.65 & 48.84 & 39.56 & 55.84 & 41.58 & 46.81 & 41.76\\
Cantor
& \textbf{49.19} & 55.20 & 51.64 & 52.02 & 46.82 & 52.71 & 47.34 & 71.79 & 50.18 & \textbf{58.15} & 51.63\\
\midrule
\noalign{\vspace{-0.5ex}}
\rowcolor{gray!8}\multicolumn{12}{c}
{\textbf{\textit{Gemma 3 27B}}}\\
\midrule
Direct
& 38.36 & 46.63 & 48.30 & 44.33 & \textbf{48.59} & \textbf{57.69} & \textbf{49.39} & 64.14 & 51.38 & 55.95 & 48.44\\
DDCoT
& 37.38 & 41.05 & 42.09 & 40.16 & 45.13 & 55.04 & 45.98 & 63.63 & 49.37 & 54.56 & 45.19\\
VoT
& 37.08 & \textbf{48.58} & \textbf{48.63} & \textbf{44.75} & 48.11 & 56.15 & 48.82 & \textbf{66.35} & \textbf{55.16} & \textbf{59.20} & \textbf{49.16}\\
VIC
& 31.97 & 30.17 & 28.43 & 30.21 & 26.01 & 31.54 & 26.50 & 26.33 & 34.40 & 31.48 & 29.17\\
Cantor
& \textbf{42.61} & 44.08 & 40.81 & 42.51 & 45.79 & 53.08 & 46.43 & 62.50 & 39.93 & 48.08 & 45.03\\

\bottomrule

\end{tabular}
}
\caption{Performance comparison of the direct prompting baseline and four MCoT methods across five base models. The best scores within each model block are highlighted in \textbf{bold}.}
\vspace{-0.5cm}
\label{tab:mcot}
\end{table*}

\subsection{Expert-Level Reasoning Results (RQ4)}
\noindent \textbf{\emph{Quantitative Reasoning (Quant.) presents greater challenges than Qualitative Reasoning (Qual.).}}
Qual. assesses directional conclusions (\eg{ ``A promotes B''}), while Quant. evaluates magnitude (\eg{ ``A stimulates B by 50\%''}). As shown in Figure~\ref{fig:RQ234}(c), models achieve higher accuracy on Observational Studies (OS), Cell Experiments (CE) and Animal Experiments (AE) than on Interventional Studies (IS) and Pathology \& Imaging (P\&I). This discrepancy arises because IS/P\&I paradigms typically require complex quantitative analysis and intricate logic, demanding stronger reasoning.

\noindent \textbf{\emph{Insights guide future AI4S research.}}
As illustrated in Figure~\ref{fig: case}, numerical perception provides foundational support, while cross-panel understanding enables experiment interpretation. Future AI4S development necessitates mature quantitative capabilities. To truly automate scientific research, MLLMs must strengthen precise, multi-step quantitative reasoning over visual evidence.

\subsection{MCoT Results (RQ5)}
\noindent \textbf{\emph{MCoT methods fail to consistently enhance performance.}}
As shown in Table~\ref{tab:mcot}, prompt-based approaches merely augment reasoning steps. Without accurately perceiving fine-grained visual content (\eg{ numerical values and cross-panel relations}), CoT guidance does not guarantee improved outcomes and can sometimes be detrimental.

\noindent \textbf{\emph{Plan-based MCoTs partially improves perception.}}
For \texttt{Qwen3-VL-30B-A3B-Instruct}, Cantor increased perception tasks accuracy from 49.50\% to 52.02\% and reasoning tasks accuracy from 57.00\% to 58.15\%. This demonstrates that predefined observation protocols can mitigate perceptual limitations, thereby enabling reasoning improvements and suggesting optimized strategies for scientific experimental image tasks.

To further decouple the effects of perception on reasoning, we selected \texttt{Qwen3-VL-30B-\allowbreak A3B-Instruct} to calculate the reasoning accuracy distribution across different MCoT methods. This analysis encompasses 423 reasoning questions, with quantitative results presented in Table~\ref{tab:mcot_rebuttal}.

\begin{table}[t]
  \centering
  \resizebox{\linewidth}{!}{
  \renewcommand{\arraystretch}{1.2}
  \begin{tabular}{l c cccc}
    \toprule
    \textbf{Condition} & \textbf{Direct} & \textbf{DDCoT} & \textbf{VoT} & \textbf{VIC} & \textbf{Cantor} \\
    \midrule
    Perception Correct & 71.66 & 82.66 & 98.59  & 65.66 & 79.65\\
    $\Delta$ vs. Direct  & --    & (\textcolor{ForestGreen}{$\uparrow 11.00$}) & (\textcolor{ForestGreen}{$\uparrow 26.93$}) & (\textcolor{Red}{$\downarrow 6.00$}) & (\textcolor{ForestGreen}{$\uparrow 7.99$}) \\
    Perception Incorrect & 32.40 & 23.68 & 9.32 & 30.30  & 40.23\\
    $\Delta$ vs. Direct  & --     & (\textcolor{Red}{$\downarrow 8.72$})  & (\textcolor{Red}{$\downarrow 23.08$}) & (\textcolor{Red}{$\downarrow 2.10$}) & (\textcolor{ForestGreen}{$\uparrow 7.83$}) \\
    \bottomrule
  \end{tabular}
  }
  \caption{Reasoning accuracy (\%) of MCoT methods decoupled by perception correctness (evaluated on \texttt{Qwen3-VL-30B-A3B-Instruct}). $\Delta$ denotes the absolute deviation from the direct prompting baseline.}
  \label{tab:mcot_rebuttal}
  \vspace{-0.2cm}
\end{table}

\noindent \textbf{\emph{MCoT significantly improves reasoning when perception is correct.}} 
Except for VIC, which ignores visual content during plan design, all MCoT methods achieved accuracy gains of 7.99\% to 26.93\% over direct prompting for the subset with correct perception. This strongly demonstrates that MCoT methods deliver substantial reasoning gains provided the MLLM initially obtains accurate visual information.

\noindent \textbf{\emph{MCoT amplifies errors when perception is incorrect.}}
Conversely, on the subset with incorrect perception, most MCoT methods yielded lower reasoning accuracy than direct prompting (declining by 2.10\% to 23.08\%). The only exception is Cantor, which intrinsically aids the model's perceptual ability. This confirms our initial observation: if models cannot accurately parse fine-grained content, additional CoT guidance amplifies perceptual hallucinations rather than correcting them.
\section{Related Work}

\paragraph{Scientific Image Reasoning}
Existing scientific image benchmarks cover diverse categories including statistical charts~\cite{ChartMoE, Vprochart, Liu2025ChartQAUnite}, schematic diagrams~\cite{Scigraphqa}, microscopy images~\cite{Microvqa}, biomedical/chemical images~\cite{laurent2024labbenchmeasuringcapabilitieslanguage,Microvqa,Micro-bench,Omnimedvqa,Chemvlm}, and general science images~\cite{Mementos,MMMU-Pro,CMMU,GRASP,feng2025sciknowevalevaluatingmultilevelscientific}. However, these benchmarks exhibit critical gaps in scientific experimental image understanding. Specifically, experimental images remain severely underrepresented in prominent benchmarks such as ScienceQA~\cite{NEURIPS2022_11332b6b}, M3CoT~\cite{M3CoT-M3}, and MMMU~\cite{Mmmu}, which predominantly feature non experimental scientific imagery. Furthermore, current benchmarks overlook cross-panel relation understanding, a fundamental aspect of experimental image interpretation. Major benchmarks including MMSci~\cite{li2024mmsci}, SciAssess~\cite{SciAssess}, SFE~\cite{zhou2025scientists}, and M3SciQA~\cite{M3SciQA} focus on isolated, low complexity images, limiting comprehensive evaluation. In contrast, \ours establishes a three-stage framework (\emph{Perception} $\rightarrow$ \emph{Understanding} $\rightarrow$ \emph{Reasoning}) with seven tasks designed to address these limitations. 

\paragraph{Multimodal Chain-of-Thought}
MCoT methods facilitate step-by-step reasoning for MLLMs in zero-shot and few-shot settings~\cite{zhang2024multimodalchainofthoughtreasoninglanguage,wang2024stop,wang2025enhancingreasoningabilitymultimodal,NEURIPS2024_62ab1c2c,Internvl,Visualization,MultiModalLatentCoT,NEURIPS2024_deeb4d6b,Video-of-Thought,SpatioTemporalVideoLang}. Existing approaches include prompt-based methods~\cite{DDCoT,VisOfThought}, plan-based decomposition~\cite{zheng2024thinkinglookingimprovingmultimodal,CantorMultimodalCoT}, and training-based frameworks~\cite{OrderChain,VLMCoTReasoning}. However, MCoT remains unverified for scientific image reasoning tasks.

\section{Conclusion}

We introduce \ours, a benchmark evaluating MLLMs' scientific experimental image \emph{perception}, \emph{understanding}, and \emph{reasoning} capabilities, alongside an assessment of various MCoT methods. Experiments reveal a significant performance gap between MLLMs and expert-level reasoning: all models except \texttt{Gemini 3 Pro Preview} fall short of the 60\% accuracy threshold. Bridging this gap, particularly in quantitative reasoning, remains a critical challenge for future research. We expect \ours to catalyze the development of more robust scientific image reasoning technologies within the AI4S community.

\section*{Limitations}

SPUR utilizes an MCQ format to evaluate MLLMs, mitigating shortcut biases through rigorous option design and filtering. However, this format provides limited visibility into the internal reasoning processes of the models. Furthermore, while SPUR is curated from open-access PMC articles across seven life science disciplines, the experimental paradigms and core challenges (\ie{ panel-level fine-grained perception, cross-panel relation understanding, and expert-level reasoning}) are representative of broader scientific contexts. Nevertheless, extending these findings to non-biological fields may require additional adaptation and dataset expansion.

\section*{Ethical Considerations}

SPUR complies with ACL ethics guidelines. This study involved no human subjects or animal experimentation. All data were collected from open-source repositories in accordance with relevant usage licenses, ensuring that privacy is preserved and no personally identifiable information is included. SPUR is released under the Creative Commons Attribution 4.0 International License (CC BY 4.0), and the associated codebase is distributed under the Apache License 2.0, supporting both commercial and open-source applications. We have made consistent efforts to minimize bias and ensure transparency throughout the dataset construction and evaluation process.

\section*{Acknowledgments}
This work is supported by the National Natural Science Foundation of China (Grant Nos. 62473271, 62176026), the Beijing Natural Science Foundation (Grant No. QY25338), and the Fundamental Research Funds for the Beijing University of Posts and Telecommunications (Grant No. 2025AI4S03). This work is also supported by the Engineering Research Center of Information Networks, Ministry of Education, China. We would also like to thank the anonymous reviewers and area chairs for constructive discussions and feedback.


\bibliography{myrefs}

\appendix


\clearpage
\section{Examples from \ours}
\label{Appdix.a}
\begin{figure}[H] 
\centering

\begin{tcolorbox}[width=\linewidth,
    colback=backblue, 
    colframe=myblue, 
    title={\centering\textcolor{white}{\textbf{Panel-Level Fine-Grained Perception: Numerical Perception}}
},  coltitle=white, 
    fonttitle=\bfseries, 
    colbacktitle=myblue, 
    enhanced, 
    width=\linewidth]

    \centering
    \includegraphics[width=0.46\linewidth]{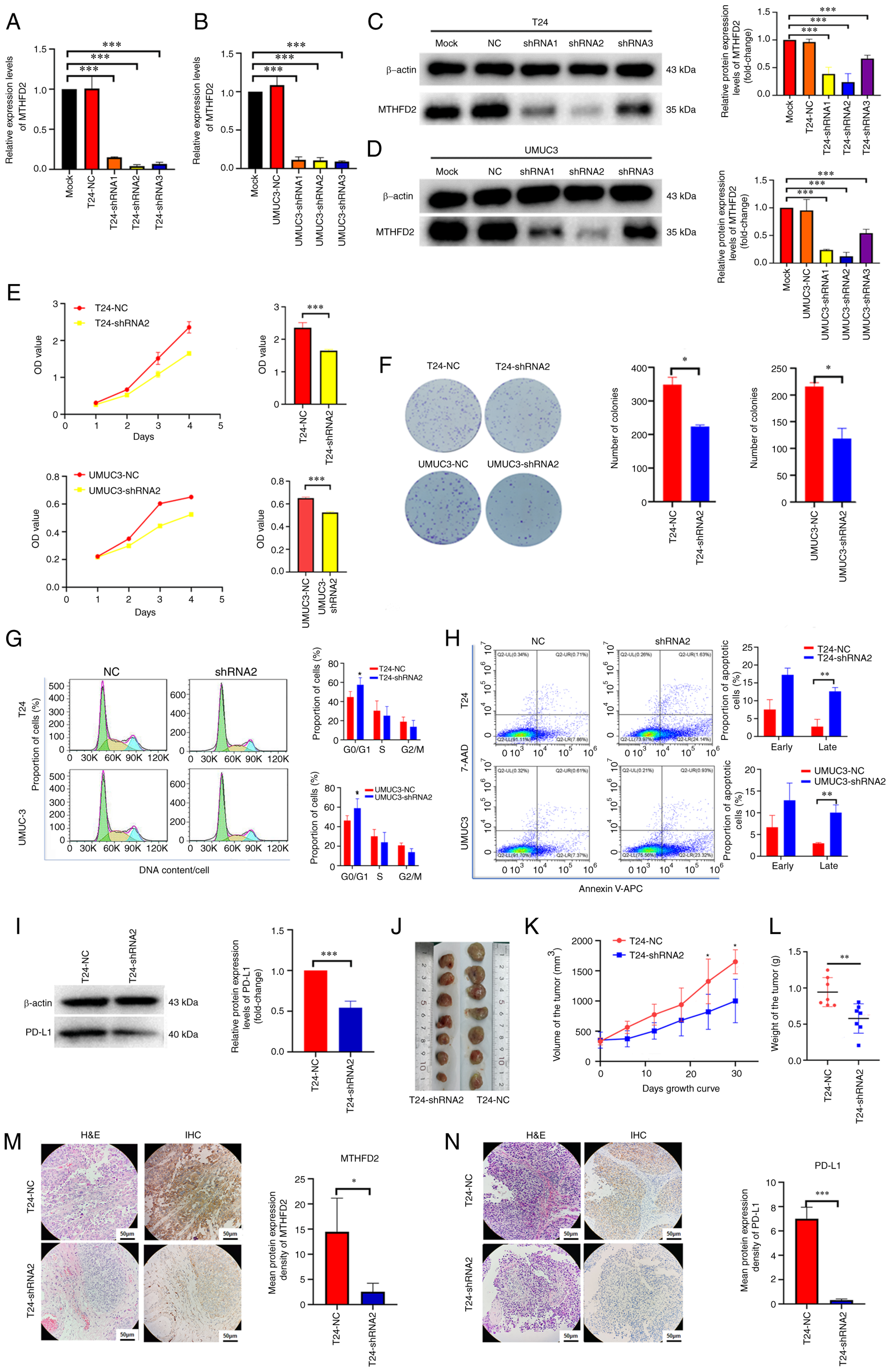}

    \raggedright  

    \vspace{1mm}
    \noindent\textbf{Question:}
    \hfill\textbf{[Perception\_2010073308\_3\_4]}\\
   Observe Fig. E Which statement is incorrect?\\
   A. In T24 cells, the proliferation curve of the shRNA2 group is consistently lower than that of the NC group.\\
B. In T24 cells, the gap between the proliferation curves of the shRNA2 group and the NC group increases over time.\\
C. The OD value of the UMUC3-shRNA2 group increases more slowly than that of the NC group.\\
D. The degrees of decrease in column height for the shRNA2 groups of both cell lines are similar.\\
E. Cannot be determined.\\

    \vspace{1mm}
    \noindent\textbf{Answer: }D \\

    \vspace{1mm}
    \noindent\textbf{Explanation:} \\
Option A: In T24 cells (Fig.~E, top left), the proliferation curve of the shRNA2 group (yellow) remains below that of the NC group (red) at all time points. Therefore, this statement is correct.\\

Option B: In T24 cells (Fig.~E, top left), the difference between the shRNA2 group and the NC group becomes more pronounced over time. For example, by Day~5, the NC group is approximately 2.2, whereas the shRNA2 group is approximately 1.4. Therefore, this statement is correct.\\

Option C: In UMUC3 cells (Fig.~E, bottom left), the OD value of the shRNA2 group increases more slowly than that of the NC group, indicating reduced proliferative activity. Therefore, this statement is correct.\\

Option D: This statement is incorrect. The reduction associated with shRNA2 is not similar between the two cell lines. Based on the plotted values, the decrease in T24 is substantially larger than that in UMUC3. Thus, the extent of reduction cannot be considered similar across the two cell lines.

\end{tcolorbox}
 
\caption{Example of a Numerical Perception (NP) task in the Panel-Level Fine-Grained Perception stage.} 

\label{fig:case1} 
\end{figure}

\begin{figure}[H] 
\centering

\begin{tcolorbox}[width=\linewidth,
    colback=backblue, 
    colframe=myblue, 
    title={\centering\textcolor{white}{\textbf{Panel-Level Fine-Grained Perception: Morphological Perception}}
},  coltitle=white, 
    fonttitle=\bfseries, 
    colbacktitle=myblue, 
    enhanced, 
    width=\linewidth]

    \centering
    \includegraphics[width=0.75\linewidth]{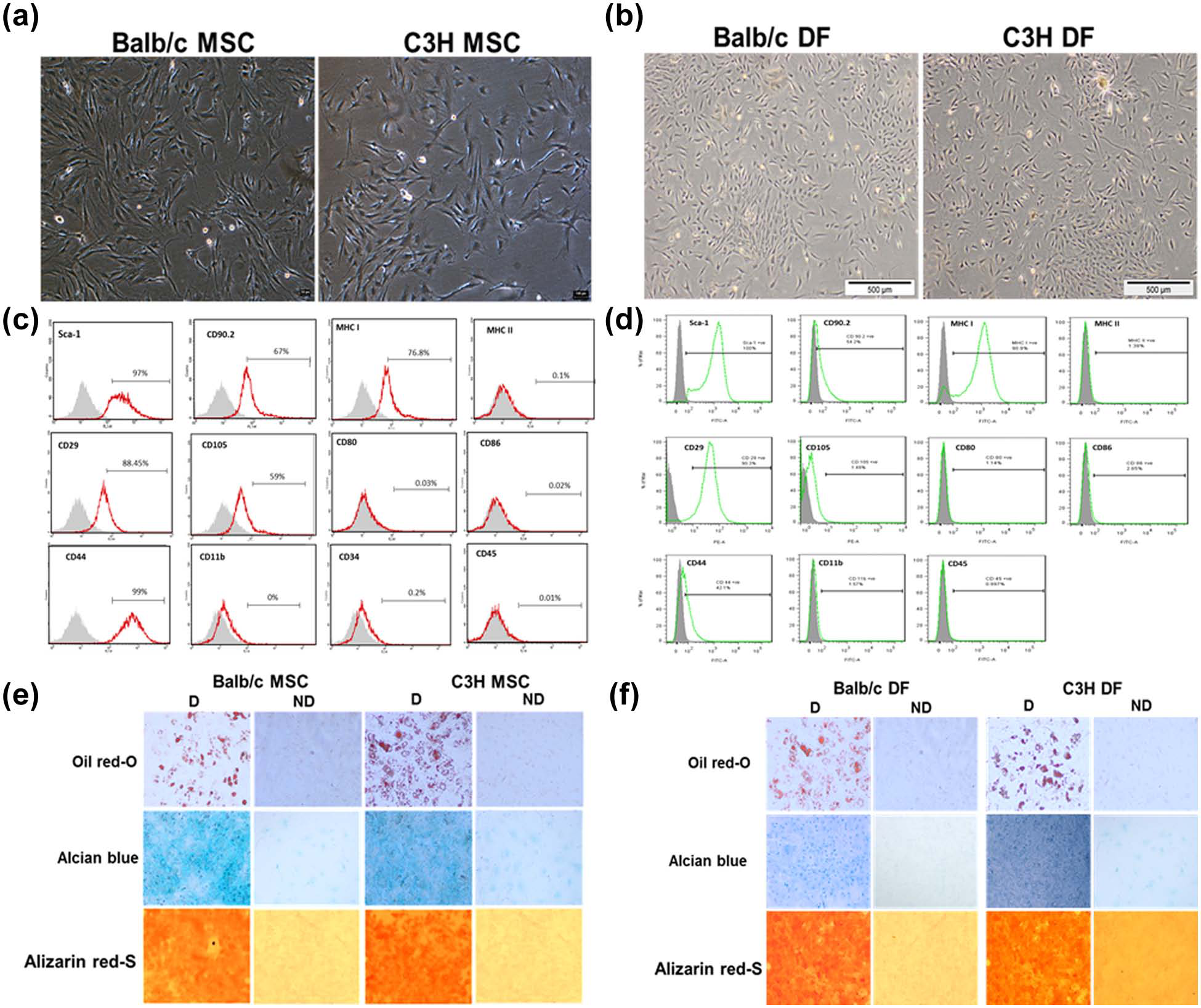}

    \raggedright  

    \vspace{1mm}
    \noindent\textbf{Question:} 
    \hfill\textbf{[Perception\_2010002387\_0\_1]}\\
   Analyze the differentiation staining results of MSC in Fig. (e) (Oil red-O, Alcian blue, Alizarin red-S; D = differentiated, ND = non-differentiated). Which statement is correct? \\
   A. In MSC Oil red-O staining, the Balb/c ND group shows more lipid droplets than the C3H D group.\\
B. Comparing Alizarin red-S staining of Balb/c MSC and C3H MSC, the orange-red mineralization signals in the differentiated groups (D) are stronger than those in the non-differentiated groups (ND) in both.\\
C. Alizarin red-S staining of MSC shows that the Balb/c D group has more obvious orange mineralization than the C3H ND group.\\
D. After induced differentiation (D group), the positive signals of the three stainings (Oil red-O, Alcian blue, Alizarin red-S) in Balb/c MSC are weaker than those in the corresponding ND groups of C3H MSC.\\
E. Cannot be determined.\\

    \vspace{1mm}
    \noindent\textbf{Answer: }B \\

    \vspace{1mm}
    \noindent\textbf{Explanation:} \\
Option A: This statement is incorrect. In Oil Red O staining, differentiated groups (D) show more lipid droplets than non-differentiated groups (ND), so the Balb/c ND cannot exceed the C3H D group.\\

Option B: In Alizarin Red S staining, both Balb/c MSC and C3H MSC show stronger mineralization signals in D than in ND. Therefore, this statement is correct.\\

Option C: Although Balb/c D appears stronger than C3H ND, this is a cross-strain comparison and is less directly supported than the within-strain D-versus-ND comparison in Option B.\\

Option D: This statement is incorrect. After differentiation, Balb/c MSC D shows stronger, not weaker, staining signals than the corresponding C3H MSC ND group.\\

Option E: This statement is incorrect. The staining images provide enough visual information to compare differentiation-related signal intensity across groups.
\end{tcolorbox}

\caption{Example of a Morphological Perception (MP) task in the Panel-Level Fine-Grained Perception stage.}
\label{fig:case2} 
\end{figure}

\begin{figure}[H] 
\centering

\begin{tcolorbox}[width=\linewidth,
    colback=backblue, 
    colframe=myblue, 
    title={\centering\textcolor{white}{\textbf{Panel-Level Fine-Grained Perception: Information Localization}}
},  coltitle=white, 
    fonttitle=\bfseries, 
    colbacktitle=myblue, 
    enhanced, 
    width=\linewidth]

    \centering
    \includegraphics[width=0.75\linewidth]{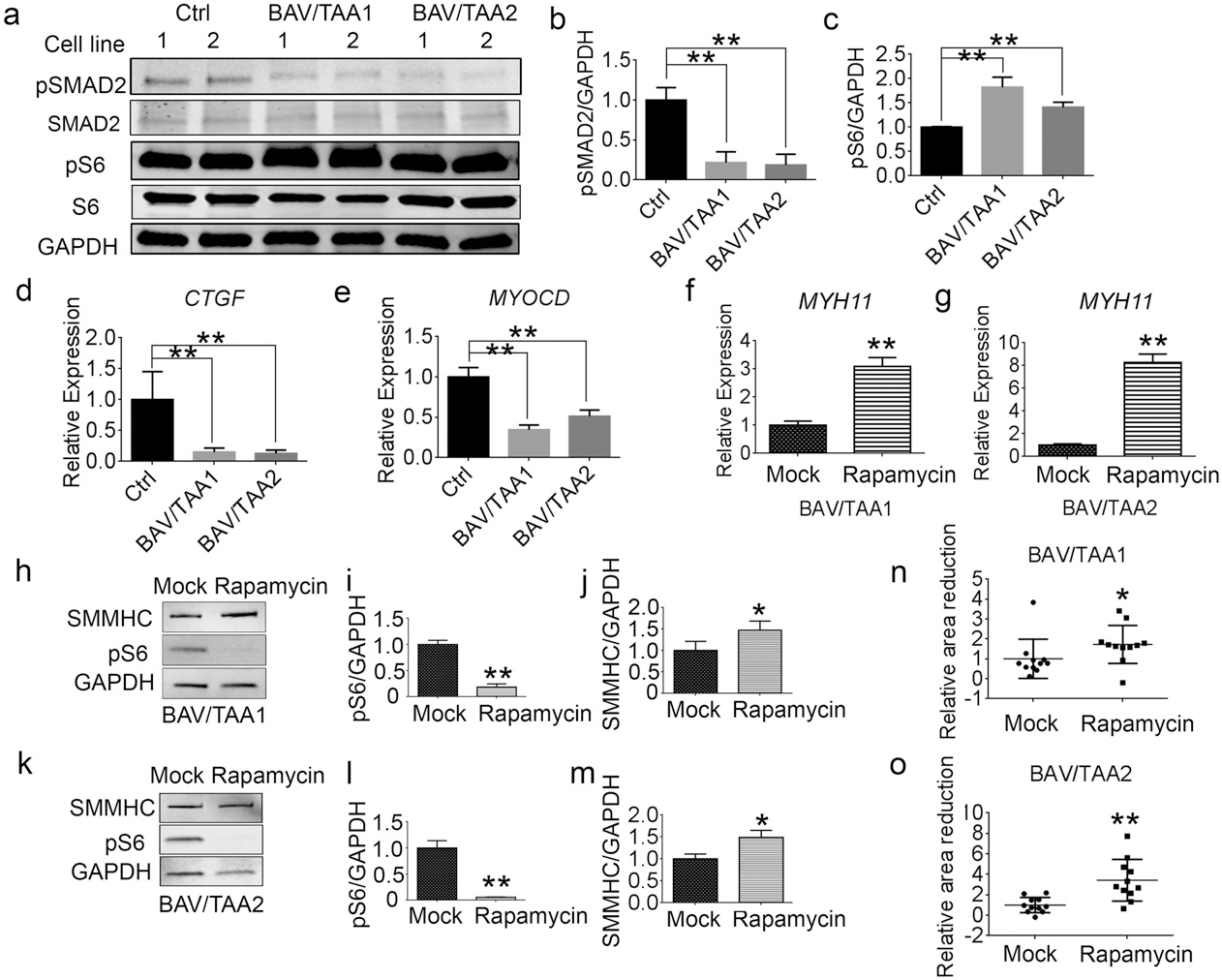}

    \raggedright  

    \vspace{1mm}
    \noindent\textbf{Question:} 
    \hfill\textbf{[Perception\_2010198904\_5\_1]}\\
   Analyze the WB bands in Fig. a (pSMAD2, SMAD2, pS6, S6, GAPDH). Which statement is correct?\\
   A. In both BAV/TAA1 and BAV/TAA2 groups, the expression of pSMAD2 is significantly higher than that in the Ctrl group.\\
B. GAPDH is stably expressed in all groups and can be used as an internal reference protein for normalizing the expression of other target proteins.\\
C. The pS6 band in the BAV/TAA2 group is darker than that in the Ctrl group.\\
D. In the Ctrl group, the expression of S6 is significantly lower than that of SMAD2.\\
E. Cannot be determined.\\

    \vspace{1mm}
    \noindent\textbf{Answer: }B \\

    \vspace{1mm}
    \noindent\textbf{Explanation:} \\
Option A: The pSMAD2 bands in the BAV/TAA1 and BAV/TAA2 groups appear weaker than those in the Ctrl group, rather than stronger. Therefore, this statement is incorrect.\\

Option B: The GAPDH bands show relatively consistent intensity across the Ctrl, BAV/TAA1, and BAV/TAA2 groups, supporting the use of GAPDH as a loading control for normalization of the target proteins. Therefore, this statement is correct.\\

Option C: The pS6 band in the BAV/TAA2 group does not appear darker than that in the Ctrl group; it appears similar in intensity. Therefore, this statement is incorrect.\\

Option D: This statement is not supported by the blot. The band intensities of S6 and SMAD2 should not be directly compared within the same lane to infer which protein is expressed at a higher level, because they are different proteins detected with different antibodies and potentially different exposure conditions. Therefore, this statement is incorrect.
    
\end{tcolorbox}
 
\caption{Example of an Information Localization (IL) task in the Panel-Level Fine-Grained Perception stage.}
\label{fig:case3} 
\end{figure}

To provide a comprehensive view of our dataset, this section visualizes representative examples across the predefined task hierarchy:

\paragraph{Panel-Level Fine-Grained Perception Cases.}
Figures~\ref{fig:case1}--\ref{fig:case3} illustrate examples of Numerical Perception (NP), Morphological Perception (MP), and Information Localization (IL).

\begin{figure}[H] 
\centering

\begin{tcolorbox}[width=\linewidth,
    colback=backblue, 
    colframe=myblue, 
    title={\centering\textcolor{white}{\textbf{Cross-Panel Relation Understanding: Trend Analysis }}
},  coltitle=white, 
    fonttitle=\bfseries, 
    colbacktitle=myblue, 
    enhanced, 
    width=\linewidth]

    \centering
    \includegraphics[width=0.44\linewidth]{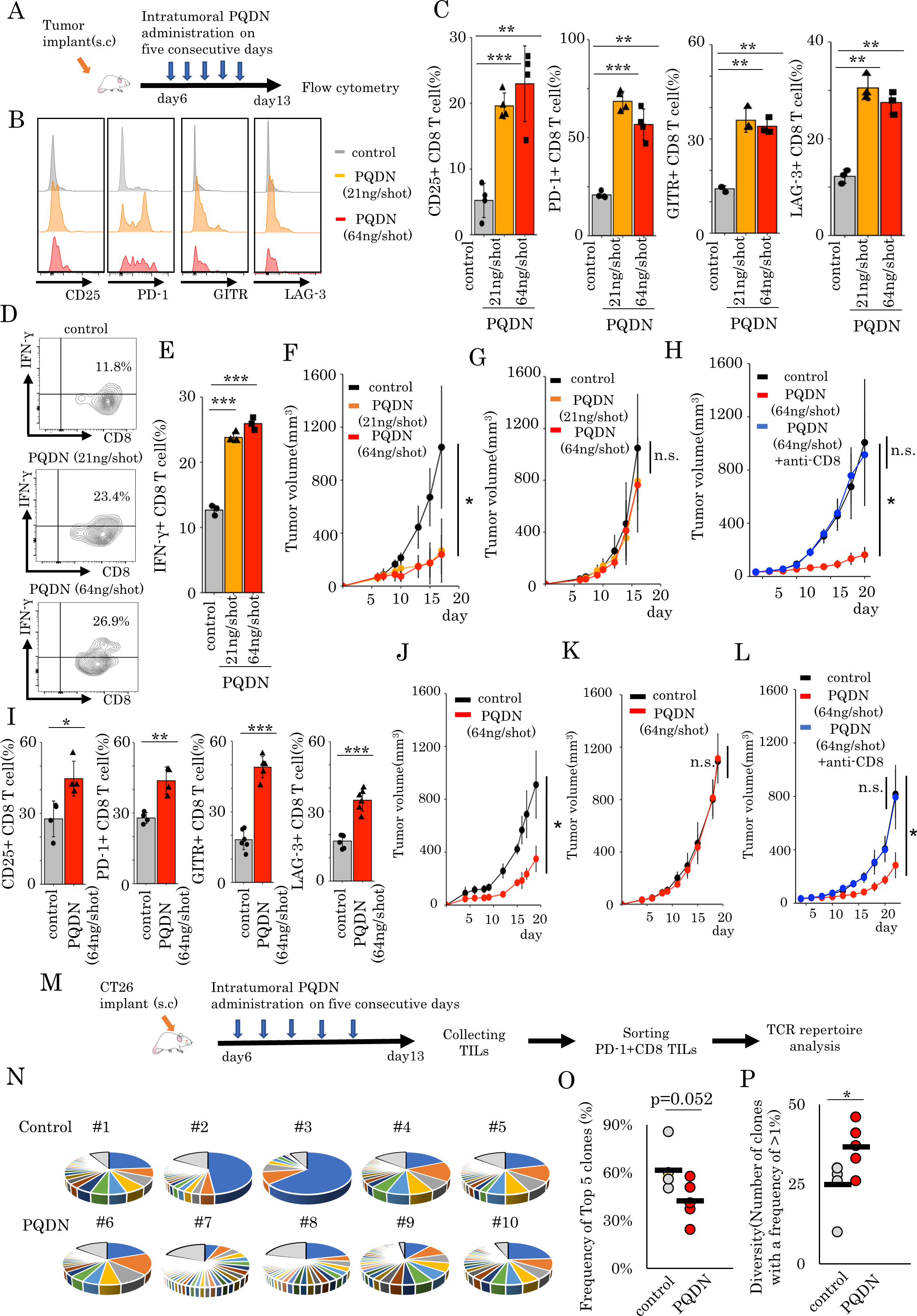}

    \raggedright  

    \vspace{1mm}
    \noindent\textbf{Question:} 
    \hfill\textbf{[Understanding\_2010105083\_3\_4]}\\
  By comparing experimental images in Fig. C, Fig. E, and Fig. F, which visual perception statement is incorrect?\\
   A. In Fig. C, PQDN groups visually show a trend of higher molecular expression with increasing dosage.\\
    B. In Fig. E, the PQDN (64ng/shot) group exhibits greater bar-height increase than the 21ng/shot group, aligning with Panel C's molecular expression trend.\\
    C. In Fig. F, tumor volume trendlines of PQDN (64ng/shot) and 21ng/shot groups overlap due to error bars in later stages, suggesting no visual difference.\\
    D. Across panels, PQDN (64ng/shot) visually demonstrates a coherent trend: rising molecular bars $\rightarrow$ elevated cytokine secretion bars $\rightarrow$ slowed tumor volume growth.\\
    E. Cannot be visually confirmed.\\
    
    \vspace{1mm}
    \noindent\textbf{Answer: }A \\

    \vspace{1mm}
    \noindent\textbf{Explanation:} \\
Option A: This statement is incorrect. In Fig. C, the PQDN groups do not show a uniformly increasing trend in molecular expression with increasing dose across all markers. Although some bars in the 64~ng/shot group appear higher than those in the 21~ng/shot group, this pattern is not sufficiently consistent to support the generalized claim of a clear dose-dependent increase.\\

Option B: In Fig. E, the PQDN 64~ng/shot group shows a greater increase in bar height than the 21~ng/shot group, and both are higher than the control group. This is visually consistent with the stronger activity observed in the higher-dose group.\\

Option C: This statement is acceptable as a visual description. In Fig. F, the tumor growth curves of the PQDN 64~ng/shot and 21~ng/shot groups become visually close in the later stages, and the overlapping error bars reduce the clarity of the difference by visual inspection alone.\\

Option D: Across Fig. C, Fig. E, and Fig. F, the PQDN 64~ng/shot group shows a broadly consistent visual pattern: relatively higher molecular-response bars, increased cytokine secretion, and slower tumor growth. Therefore, this statement is supported at the level of overall visual trend.
    
\end{tcolorbox}

\caption{Example of a Trend Analysis (TA) task in the Cross-Panel Relation Understanding stage.}
\label{fig:case4} 
\end{figure}

\begin{figure}[H] 
\centering

\begin{tcolorbox}[width=\linewidth,
    colback=backblue, 
    colframe=myblue, 
    title={\centering\textcolor{white}{\textbf{Cross-Panel Relation Understanding: Heterogeneous Integration }}
},  coltitle=white, 
    fonttitle=\bfseries, 
    colbacktitle=myblue, 
    enhanced, 
    width=\linewidth]

    \centering
    \includegraphics[width=0.39\linewidth]{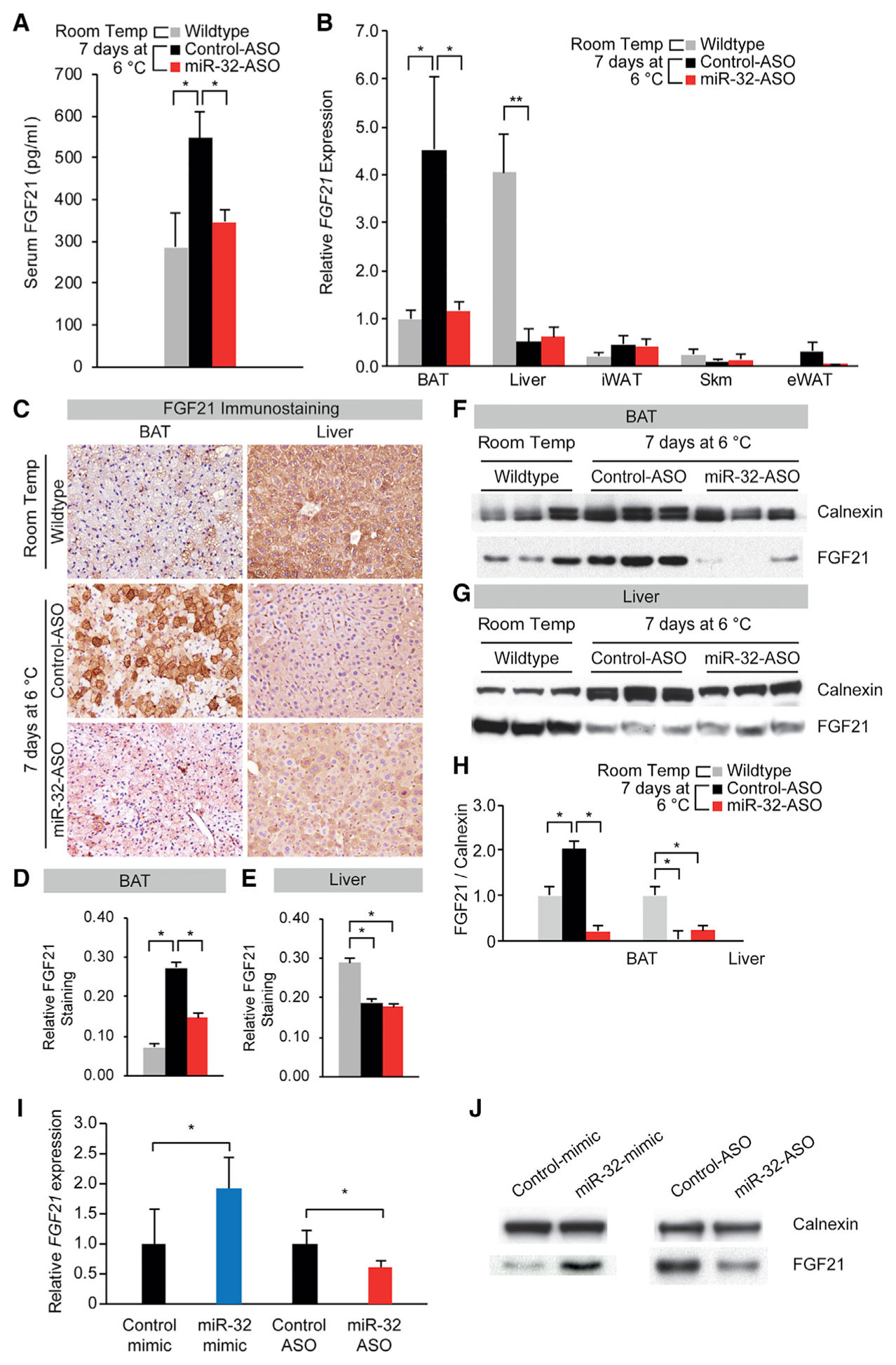}

    \raggedright  

    \vspace{1mm}
    \noindent\textbf{Question:} 
    \hfill\textbf{[Understanding\_2010143105\_4\_1]}\\
   How do the visual FGF21 expression levels correlate between the staining and protein band results?\\
   A. Both Fig. C and Fig. F visually show significant FGF21 enhancement by cold stimulation, which is then substantially suppressed by miR-32-ASO treatment in both, demonstrating a highly consistent trend.\\
    B. Fig. C's FGF21 staining enhances with cold but visually shows no significant change after miR-32-ASO treatment, while Fig. F's FGF21 protein band shows enhancement followed by suppression. Their visual trends are not entirely consistent.\\
    C. Under cold stimulation, Fig. C's FGF21 staining visually weakens, contrasting with the visual enhancement trend of Fig. F's FGF21 protein band.\\
    D. miR-32-ASO treatment visually enhances FGF21 staining in Fig. C, and simultaneously, the FGF21 protein band in Fig. F also visually strengthens.\\
    E. Cannot be determined.\\

    \vspace{1mm}
    \noindent\textbf{Answer: }A \\

    \vspace{1mm}
    \noindent\textbf{Explanation:} \\
Option A: Both Fig. C (immunostaining) and Fig. F (Western blot) show consistent trends. Under cold stimulation (7 days at 6~$^\circ$C, Control-ASO group), Fig. C demonstrates darkened BAT/Liver FGF21 staining (enhancement), and Fig. F demonstrates thickened/brightened BAT FGF21 bands (enhancement). After miR-32-ASO treatment, Fig. C shows lighter staining (suppression), and Fig. F shows weakened/disappeared bands (suppression). Thus, trends align perfectly across methods.\\
Option B: This option claims that Fig. C shows no significant change after miR-32-ASO treatment. However, Fig. C clearly exhibits lighter staining, directly contradicting this claim. Therefore, it is incorrect.\\
Option C: This option claims that cold weakens Fig. C staining, but Fig. C actually shows darkening (enhancement) with cold, which is consistent with Fig. F. Therefore, it is incorrect.\\
Option D: This option claims that miR-32-ASO enhances FGF21, but both methods clearly demonstrate weaker signals (suppression). Therefore, it is incorrect.
    
\end{tcolorbox}

\caption{Example of a Heterogeneous Integration (HI) task in the Cross-Panel Relation Understanding stage.}
\label{fig:case5} 
\end{figure}

\paragraph{Cross-Panel Relation Understanding Cases.} Figures~\ref{fig:case4} and~\ref{fig:case5} present examples of Trend Analysis (TA) and Heterogeneous Integration (HI).

\begin{figure}[H] 
\centering

\begin{tcolorbox}[width=\linewidth,
    colback=backblue, 
    colframe=myblue, 
    title={\centering\textcolor{white}{\textbf{Qualitative Reasoning: Chart + Staining Image + Western Blot}}
},  coltitle=white, 
    fonttitle=\bfseries, 
    colbacktitle=myblue, 
    enhanced, 
    width=\linewidth]

    \centering
    \includegraphics[width=0.75\linewidth]{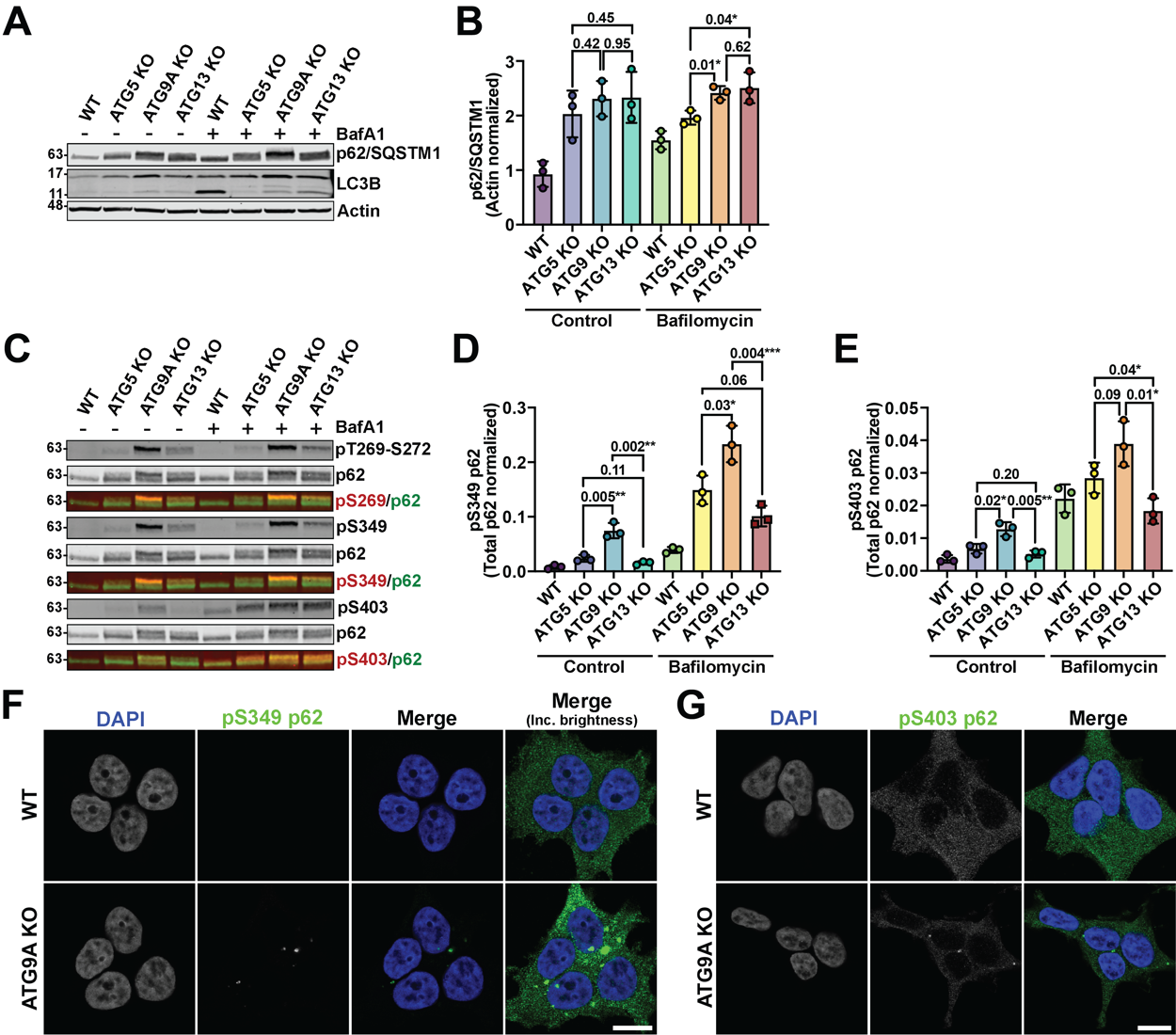}

    \raggedright  

    \vspace{1mm}
    \noindent\textbf{Question:} 
    \hfill\textbf{[Reasoning\_2010143347\_3\_2]}\\
   Based on the experimental images, which of the following is correct?\\
   A. In Fig. A, ATG5 knockout leads to p62 accumulation, indicating blocked autophagic degradation independent of other ATG genes.\\
B. In Fig. B, p62 levels remain unchanged across ATG knockouts after BafA1 treatment, suggesting functional similarity.\\
C. Figs. C--E show phosphorylated p62 in ATG9A knockout, indicating its role in regulating p62 phosphorylation.\\
D. Figs. F and G show ATG9A knockout does not affect p62 localization, suggesting no impact on subcellular distribution.\\
E. Cannot be determined.\\

    \vspace{1mm}
    \noindent\textbf{Answer: }C \\

    \vspace{1mm}
    \noindent\textbf{Explanation:} \\
Option A: In Fig.~A, ATG5 knockout leads to p62 accumulation, but it cannot indicate that the blocked autophagic degradation is independent of other ATG genes, because there may be synergistic interactions among autophagy-related genes. The claim of ``independence'' goes beyond the scope of the data. Thus, A is incorrect.\\
Option B: In Fig.~B, the p62 levels of different ATG gene knockout groups change differently after BafA1 treatment, indicating that these ATG genes have different functions in regulating autophagy, rather than similar ones. Thus, B is incorrect.\\
Option C: Figs.~C--E show that in the ATG9A knockout group, the phosphorylation form of p62 changes, which indicates that ATG9A is involved in regulating the phosphorylation of p62. Thus, C is correct.\\
Option D: Figs.~F and G show that the subcellular localization of p62 changes in the ATG9A knockout group, indicating that ATG9A affects the subcellular distribution of p62. Thus, D is incorrect.
    
\end{tcolorbox}

\caption{Example of a Qualitative Reasoning task in the Expert-Level Reasoning stage.} 
\label{fig:case6} 
\end{figure}

\begin{figure}[H] 
\centering

\begin{tcolorbox}[width=\linewidth,
    colback=backblue, 
    colframe=myblue, 
    title={\centering\textcolor{white}{\textbf{Quantitative Reasoning: Chart + Staining Image + Western Blot}}
},  coltitle=white, 
    fonttitle=\bfseries, 
    colbacktitle=myblue, 
    enhanced, 
    width=\linewidth]

    \centering
    \includegraphics[width=0.6\linewidth]{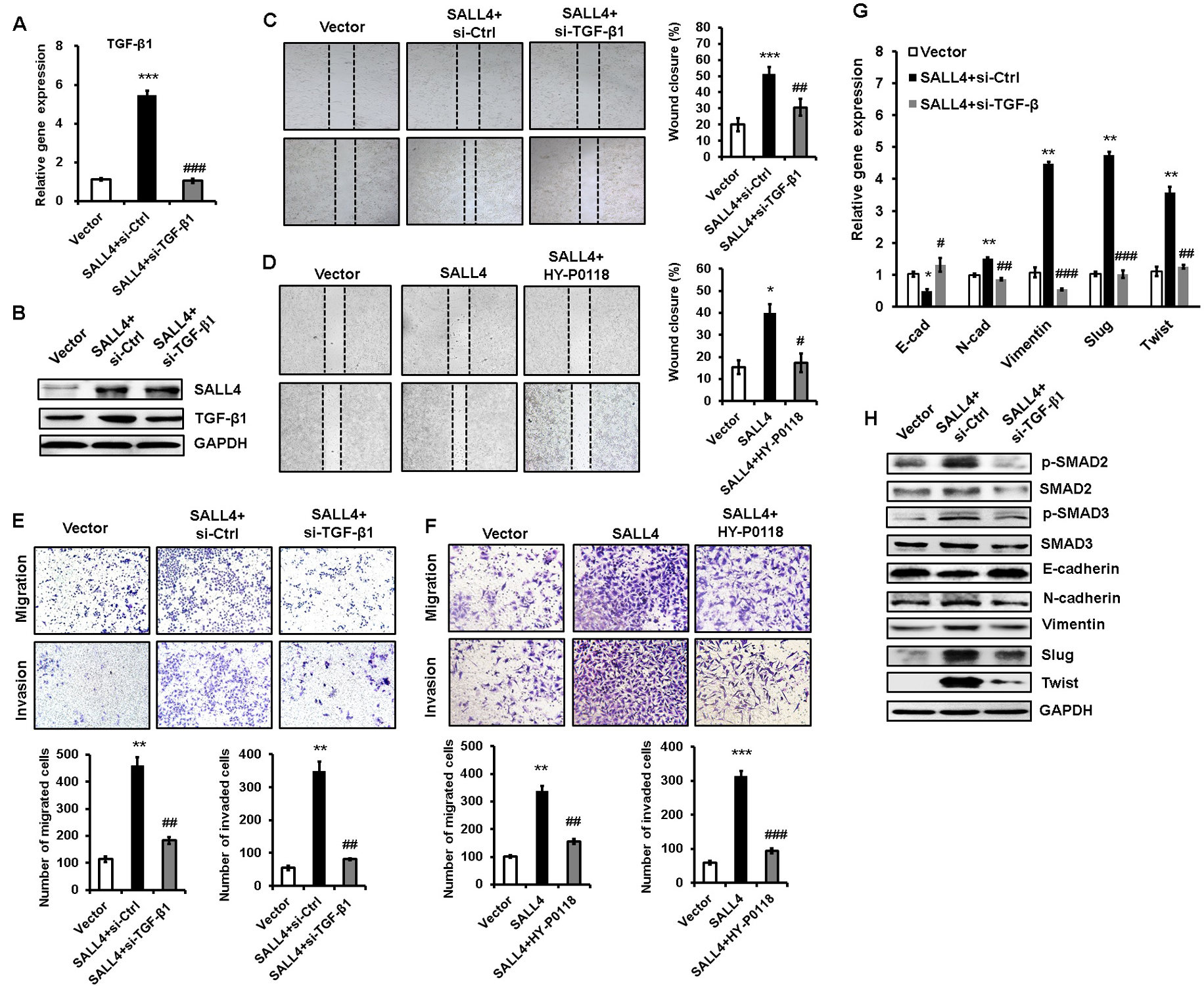}

    \raggedright  

    \vspace{1mm}
    \noindent\textbf{Question:} 
    \hfill\textbf{[Reasoning\_2010045119\_4\_2]}\\
   In SALL4-overexpressing HGC-27 cells, which best describes key changes?\\
   A. SALL4 raises TGF-$\beta$1 mRNA 6$\times$, wound closure $\sim$ 60\%, invasion 5$\times$; si-TGF-$\beta$1 cuts migration $>$ 70\%.\\
B. si-TGF-$\beta$1 lowers Slug, Vimentin, Twist mRNA 60\%--80\%, protein and p-SMAD2/3 decrease, showing TGF-$\beta$1/SMAD inhibition.\\
C. HY-P0118 reduces wound closure 60\% $\rightarrow$ 20\%, migration 450 $\rightarrow$ 100, p-SMAD2/3 returns to control, blocking TGF-$\beta$1.\\
D. si-TGF-$\beta$1 leaves N-cadherin, Vimentin mRNA unchanged, minor protein drop, E-cadherin stable; no EMT reversal.\\
E. Cannot be determined.\\

    \vspace{1mm}
    \noindent\textbf{Answer: }B \\

    \vspace{1mm}
    \noindent\textbf{Explanation:} \\
Option A: This statement is not the best description. Although SALL4 overexpression appears to increase TGF-$\beta$1 expression, wound closure, and invasion, the numerical claims (such as ``6$\times$,'' ``$\sim$ 60\%,'' ``5$\times$,'' and ``$>$ 70\%'') are overly specific and combine results from different assays into a single summary that is not the most directly supported by the figure. Therefore, A is not the correct choice.

Option B: In Figs.~G and H, si-TGF-$\beta$1 reduces the mRNA levels of Slug, Vimentin, and Twist, and is also associated with decreased protein levels and reduced p-SMAD2/3 signals. These findings support inhibition of the TGF-$\beta$1/SMAD pathway and partial reversal of the EMT-related phenotype. Therefore, B is the best-supported statement.

Option C: Although HY-P0118 appears to reduce wound closure, migration/invasion, and p-SMAD2/3 levels, the statement that p-SMAD2/3 ``returns to control'' is too strong and not directly established by the figure. Therefore, C is incorrect.

Option D: This statement is contradicted by the data. After si-TGF-$\beta$1 treatment, N-cadherin and Vimentin decrease rather than remain unchanged, and E-cadherin increases rather than stays stable, indicating reversal rather than persistence of the EMT-related phenotype. Therefore, D is incorrect.

Option E: The figure provides clear visual and quantitative evidence for changes in TGF-$\beta$1 expression, cell migration/invasion, and EMT-related markers. Therefore, the results can be determined from the data.
    
\end{tcolorbox}

\caption{Example of a Quantitative Reasoning task in the Expert-Level Reasoning stage.}
\label{fig:case7} 
\end{figure}

\paragraph{Expert-Level Reasoning Cases.} Figures~\ref{fig:case6} and~\ref{fig:case7} demonstrate Qualitative Reasoning (Qual.) and Quantitative Reasoning (Quant.) featuring a representative multi-panel configuration.

\section{Details of Annotation}
\label{Appdix.b}
This section provides details on the construction and annotation pipeline of \ours. Table~\ref{tab:Annotation_Statistics} outlines the precise quantitative breakdown of data retention and exclusion at each step.

\begin{table*}[!ht]
  \centering
  \resizebox{1.0\linewidth}{!}{
  \renewcommand{\arraystretch}{1.1}
  \begin{tabular}{l l c c}
        \toprule
        Screening Step       & Operation Description  & Removed Quantity & Retained Quantity \\
        \midrule
        Initial Image Extraction    & Extracted candidate images from PMC papers        & --                & Images: 5,632 \\
        Complexity Filtering   & Excluded images with $\leq$ 6 panels via YOLO detector    & Images: 4,368    & Images: 1,264 \\
        Raw Image Review & Excluded images lacking complete workflows or methodological rigor     & Images: 180      & Images: 1,084 \\
        QA Pair Generation     & Generated candidate QA pairs using GPT-4o based on retained images     & --    & Images: 1,084; QA Pairs: 7,608 \\
        Textual Shortcut Elimination  & Discarded QA pairs with $\geq$ 5 correct text-only responses & QA Pairs: 1,612  & Images: 1,084; QA Pairs: 5,996 \\
        QA Pair Review   & Discarded QA pairs with factual errors or task misalignment     & QA Pairs: 1,732  & Images: 1,084; QA Pairs: 4,264 \\
        \bottomrule
  \end{tabular}
  }
  \caption{Quantitative breakdown of data retention and exclusion at each curation step of \ours.}  
  \label{tab:Annotation_Statistics}
\end{table*}

\subsection{Image Collection and Preprocessing}

\begin{figure*}[!ht]
	\centering
	\includegraphics[width=0.96\textwidth]{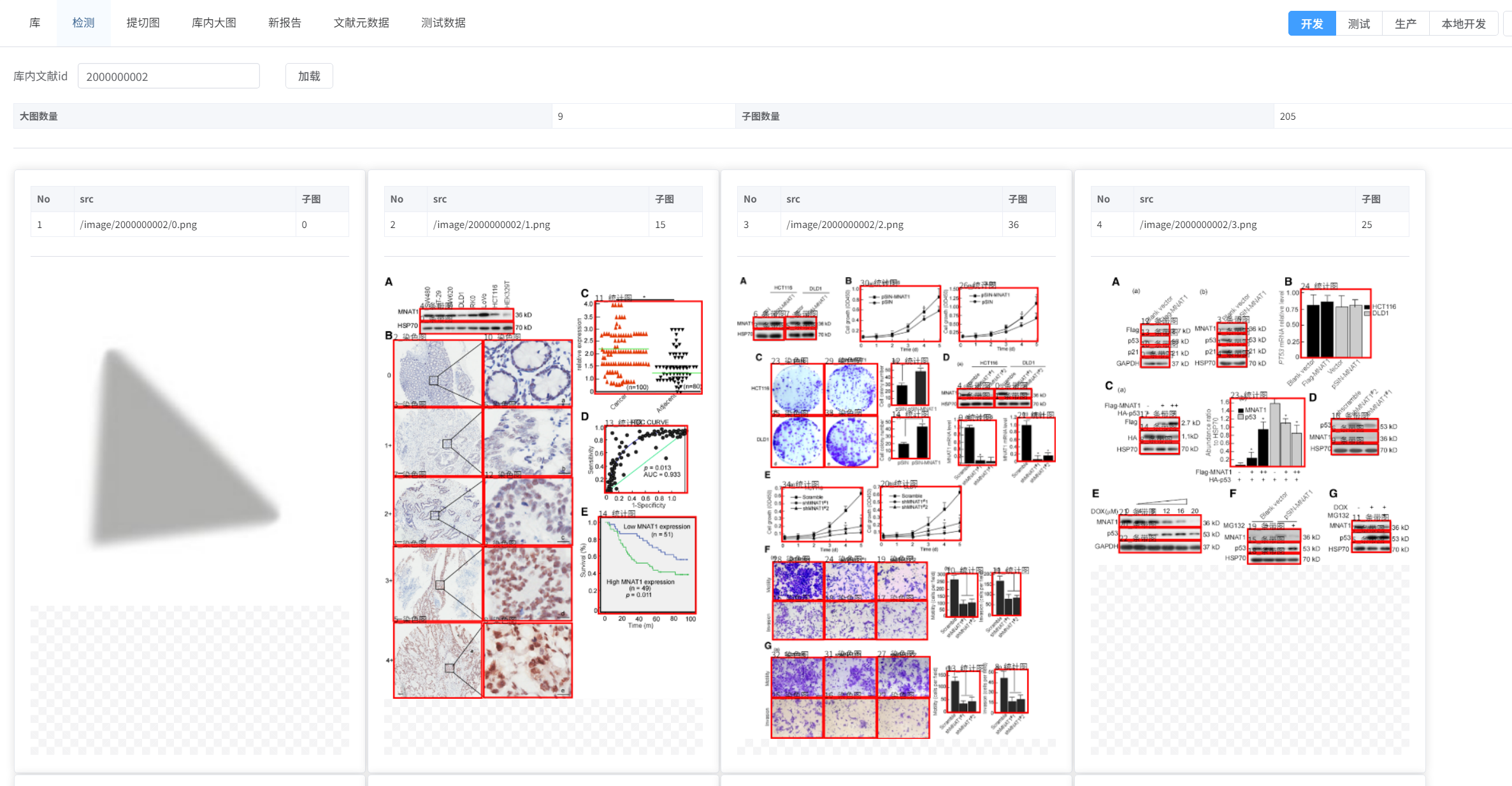}
	\caption{User interface of the academic paper selection platform utilized during the paper-level curation step.}
	\label{appendix.page}
\end{figure*}

\paragraph{Source Paper Selection Criteria.}
Candidate papers were strictly limited to open-access PMC publications from the past 10 years with a source journal Impact Factor (IF) $>$ 3.0, ensuring both scientific timeliness and authority. Human annotators confirmed the presence of complex experimental images, explicitly excluding review papers, meta-analyses, and theoretical studies lacking original experimental data. Figure~\ref{appendix.page} illustrates the user interface of the selection platform.

\paragraph{Manual Image--Text Alignment.} For the retained images, annotators performed the following standardization: (1) extracted 3--5 key sentences from the main text detailing the experimental background, methods, or results; (2) standardized the original captions by correcting typos, supplementing ambiguous acronyms, and ensuring terminology consistency; and (3) categorized the images into seven predefined disciplinary categories (\eg{ Cell Biology, Molecular Biology, and Oncology}) based on the source journal.

\subsection{Panel Recognition}

\begin{figure*}[t!]
	\centering
	\includegraphics[width=0.98\textwidth]{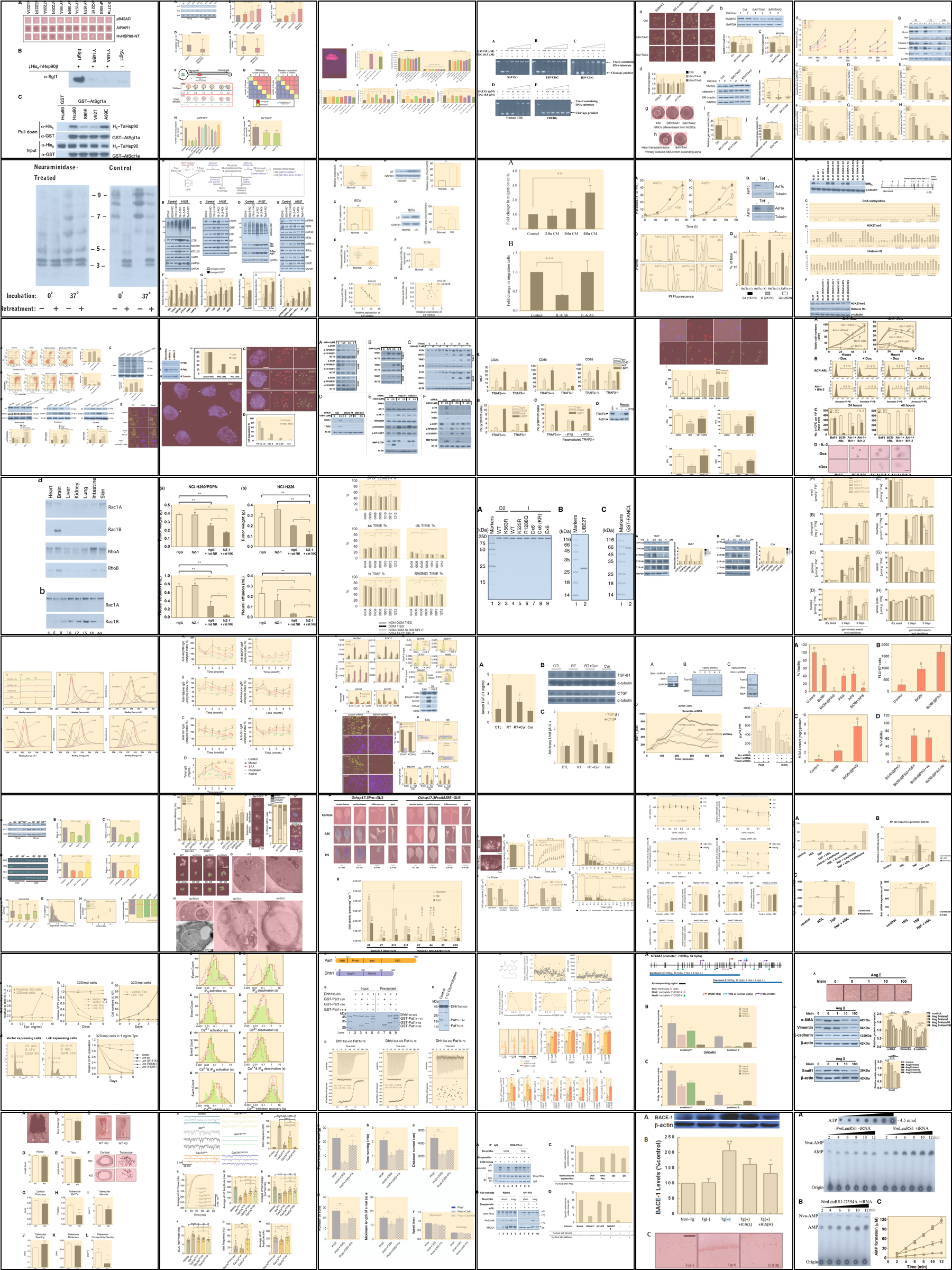}
	\caption{Visualization of panel category detections in \ours. \textcolor{mask_red}{Red}, \textcolor{mask_yellow}{light yellow}, and \textcolor{mask_blue}{light blue} masks denote staining images (comprising four fine-grained subtypes), charts, and Western blots, respectively.}
	\label{appendix.1}
\end{figure*}

The YOLO-based panel detector for the initial complexity filter was pre-trained on a proprietary dataset comprising annotated panel coordinates and categories from biomedical and materials science journals. This model automatically excluded images with $\leq$ 6 panels to guarantee structural complexity. Figure~\ref{appendix.1} visualizes the detection of the six recognized panel categories in the final dataset.

\subsection{Expert Validation Protocols}
\label{appendix.expert}

To ensure methodological rigor and scientific validity, the retained data underwent multi-tiered expert reviews.

\paragraph{Raw Image Review.}

\begin{itemize} [leftmargin=*]
    \item \textbf{Validation Criteria.} Reviewers verified experimental design completeness (\eg{ clear control groups, experimental variables, and measurable outcomes}), methodological integrity (\eg{ clear sample labeling, appropriate reagents, and repeatable operations}), and image representativeness.
    \item \textbf{Execution Rules.} Each image was independently evaluated by two experts. Images were retained upon dual approval and excluded upon dual rejection. Conflicting judgments underwent decisive arbitration by a senior expert. Initial consistency was reached on 78\% of the images, with 22\% requiring senior arbitration.
\end{itemize}

\paragraph{QA Pair Review.} Retained questions were evaluated along three dimensions (scored 0--1) to eliminate factual inaccuracies or multimodal misalignments:

\begin{itemize} [leftmargin=*]
    \item \textbf{Scientific Validity.} Conformance to scientific facts, absence of logical errors in conclusions, and correct usage of professional terminology.
    
    \item \textbf{Task Alignment.} Verification that the QA pair strictly aligns with the predefined task hierarchy and that the required reasoning matches the visual data (\eg{ a ``quantitative trend analysis'' query must correspond to quantifiable chart data}).
    
    \item \textbf{Visual Reasoning Necessity.} Final confirmation that the question cannot be answered via textual shortcuts or common sense, strictly requiring visual interpretation of the image.
\end{itemize}

\subsection{Fine-Grained Distribution}
\label{appendix.diversity}
\begin{figure*}[t!]
	\centering
	\includegraphics[width=0.6\textwidth]{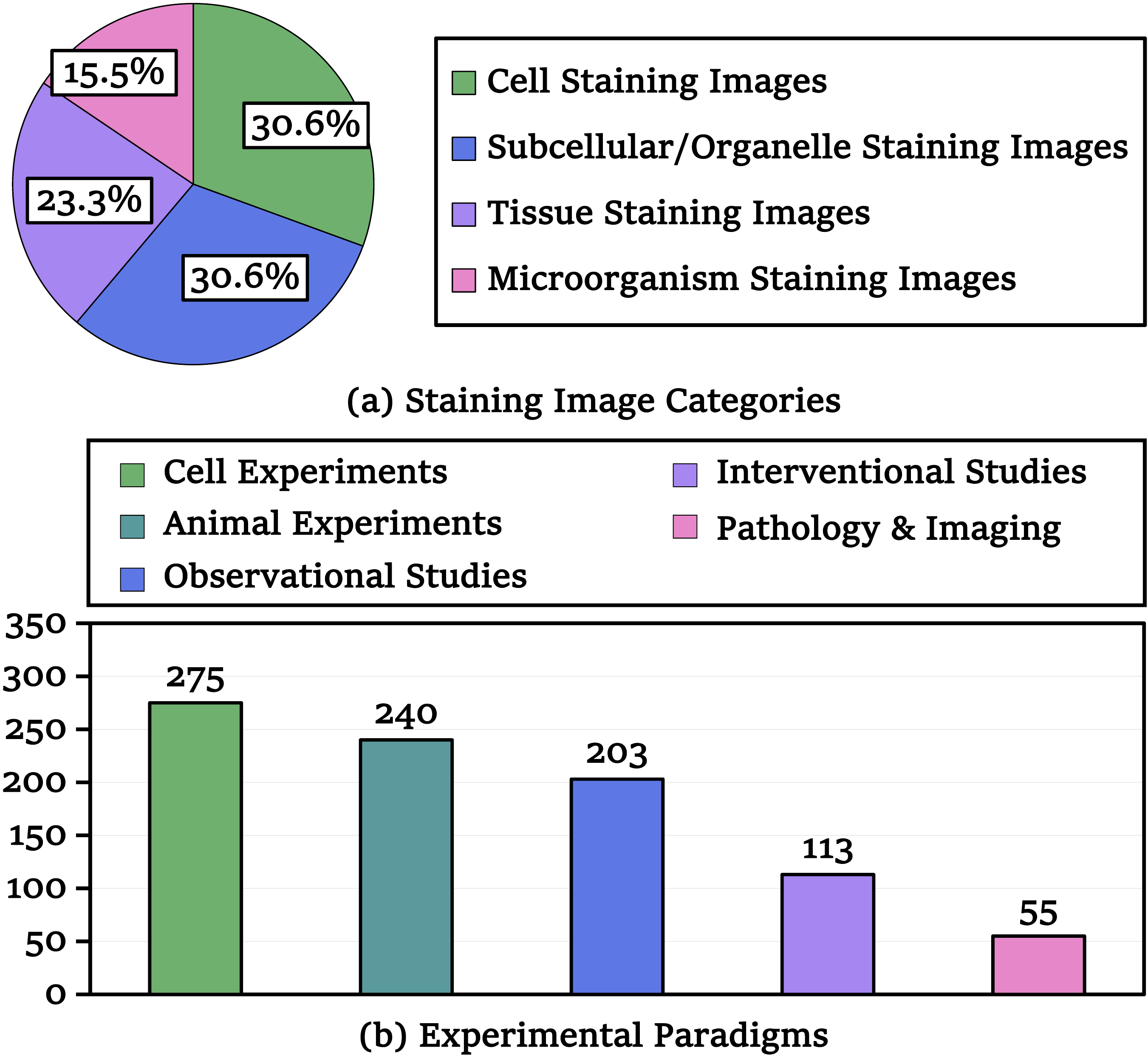}
    \caption{Distribution of \textbf{(a)} Staining Image Categories and \textbf{(b)} Experimental Paradigms in \ours.}
	\label{fig: Diversity}
\end{figure*}

As illustrated in Figure~\ref{fig: Diversity}, \ours further categorizes staining images into four fine-grained subtypes (\eg{ Cell and Tissue}) within the Panel-Level Fine-Grained Perception stage, and evaluates performance across five typical experimental paradigms within the Expert-Level Reasoning stage.

\section{Experiment Details and Analysis}
\label{Appdix.c}

\subsection{Evaluated Models}
\label{Appdix.c.1}

Since \ours requires visual interpretation of complex scientific figures, we exclusively evaluate MLLMs with advanced vision capabilities. Table~\ref{app:Tab.4} details the proprietary and open-source models evaluated, including their providers, release dates, versions, and parameter sizes.

\subsection{Fine-Grained Analysis}
\label{Appdix.c.2}

\paragraph{Staining Image Categories.} This analysis evaluates MLLM performance on fine-grained biological structures within the Panel-Level Fine-Grained Perception stage, specifically categorizing staining images into Cell, Tissue, Microorganism, and Subcellular/Organelle. Detailed quantitative results are presented in Table~\ref{app:Tab.5}.

\paragraph{Number of Relations.} Within the Cross-Panel Relation Understanding stage, task complexity is quantified by the number of cross-panel relation pairs examined across all options of a given question.

\paragraph{Experimental Paradigms.} We further analyze model capabilities within the Expert-Level Reasoning stage across five fundamental scientific experimental paradigms: Cell Experiments (CE), Animal Experiments (AE), Observational Studies (OS), Interventional Studies (IS), and Pathology \& Imaging (P\&I). Performance breakdown in these Expert-Level Reasoning tasks is detailed in Table~\ref{app:Tab.6}.

\section{Use of AI Assistants}
\label{sec:ai_assistants}

This research was driven entirely by the authors, who provided all core scientific insights, experimental designs, and analyses. We acknowledge the use of AI assistants during the preparation of this manuscript: Cursor was utilized to aid in code writing and data processing, while large language models (LLMs) assisted with language editing to enhance readability. Furthermore, \texttt{GPT-4o} assisted in QA pair construction for our benchmark, with detailed methodologies in Appendix~\ref{Appdix.b}. We emphasize that all AI-assisted content was thoroughly reviewed and validated by the authors, who bear full responsibility for the scientific integrity and fundamental contributions of this work.


\begin{table*}[!t]
\centering
\resizebox{\textwidth}{!}{%
\renewcommand{\arraystretch}{1.1}

\begin{tabular}{llclc}
\toprule
\textbf{Provider}    & \textbf{Model}           & \textbf{Release} & \textbf{Version}                    & \textbf{Size} \\ 
\midrule
\noalign{\vspace{-0.5ex}}
\rowcolor{gray!8}\multicolumn{5}{c}
{\textbf{\textit{Proprietary MLLMs}}}\\
\midrule

\multirow{3}{*}{OpenAI}
& GPT-5.1 & 2025-11 & gpt-5.1 & -- \\
& OpenAI o4-mini-high & 2025-04 & o4-mini-high & -- \\
& GPT-4o  & 2024-11 & gpt-4o-2024-11-20 & -- \\
\noalign{\vskip 0.5ex}\hdashline\noalign{\vskip 0.5ex}
\multirow{2}{*}{Google}
& Gemini 3 Pro Preview& 2025-11 & gemini-3-pro-preview & -- \\
& Gemini 2.5 Pro Preview& 2025-06 & gemini-2.5-pro-preview-06-05 & -- \\
\noalign{\vskip 0.5ex}\hdashline\noalign{\vskip 0.5ex}
Anthropic                
& Claude 3.7 Sonnet (thinking) & 2025-02 & claude-3.7-sonnet:thinking & -- \\
\noalign{\vskip 0.5ex}\hdashline\noalign{\vskip 0.5ex}
ByteDance                
& Doubao-Seed-1.6 & 2025-06 & doubao-seed-1.6-250615 & -- \\
\noalign{\vskip 0.5ex}\hdashline\noalign{\vskip 0.5ex}
xAI
& Grok 4.1 Fast & 2025-11 & grok-4.1-fast & -- \\
\midrule

\noalign{\vspace{-0.5ex}}
\rowcolor{gray!8}\multicolumn{5}{c}
{\textbf{\textit{Open-Source MLLMs}}}\\
\midrule

Meta  
& Llama 4 Maverick & 2025-04 & llama-4-maverick & 400B-A17B \\
\noalign{\vskip 0.5ex}\hdashline\noalign{\vskip 0.5ex}
Google                   
& Gemma 3 27B & 2025-03 & gemma-3-27b-it  & 27B \\
\noalign{\vskip 0.5ex}\hdashline\noalign{\vskip 0.5ex}
\multirow{3}{*}{Mistral AI}
& Ministral 3 14B& 2025-12 & ministral-14b-2512 & 14B \\
& Ministral 3 8B & 2025-12 & ministral-8b-2512 & 8B \\
& Mistral Small 3.1& 2025-03 & mistral-small-3.1-24b-instruct-2503 & 24B \\
\noalign{\vskip 0.5ex}\hdashline\noalign{\vskip 0.5ex}
\multirow{3}{*}{Alibaba} 
&  Qwen3-VL-30B-A3B-Thinking  & 2025-10 & qwen3-vl-30b-a3b-thinking & 30B-A3B \\
& Qwen3-VL-30B-A3B-Instruct  & 2025-10 & qwen3-vl-30b-a3b-instruct & 30B-A3B \\
& Qwen2.5-VL-72B  & 2025-01 & qwen2.5-vl-72b-instruct & 72B \\
\noalign{\vskip 0.5ex}\hdashline\noalign{\vskip 0.5ex}
OpenGVLab            
& InternVL3-78B  & 2025-04 & internvl3-78b & 78B \\
\noalign{\vskip 0.5ex}\hdashline\noalign{\vskip 0.5ex}
\multirow{2}{*} {LLaVA Community}   
& LLaVA-OneVision-7B & 2024-08  & llava-onevision-qwen2-7b-ov-hf & 7B \\
& LLaVA-v1.5-13B & 2023-09 & llava-v1.5-13b-hf & 13B \\
\noalign{\vskip 0.5ex}\hdashline\noalign{\vskip 0.5ex}
Z.ai
& GLM-4.5V & 2025-08 & glm-4.5v & 106B-A12B \\

\bottomrule
\end{tabular}
}
\caption{Detailed specifications of the proprietary and open-source MLLMs evaluated on \ours.}
\label{app:Tab.4}
\end{table*}

\begin{table*}[t]
\centering
\resizebox{\textwidth}{!}{
\begin{tabular}{l c c c c c}
\toprule
\textbf{Model} & \textbf{Micro Avg.} & \textbf{Cell} & \textbf{Tissue} & \textbf{Microorganism} & \textbf{\makecell[c]{Subcellular/\\Organelle}} \\
\midrule
\noalign{\vspace{-0.5ex}}
\rowcolor{gray!8}\multicolumn{6}{c}
{\textbf{\textit{Proprietary MLLMs}}}\\
\midrule

Gemini 3 Pro Preview& 62.92 & 66.92\,(\textcolor{ForestGreen}{$\uparrow 4.00$}) & 61.14\,(\textcolor{Red}{$\downarrow 1.78$})& 53.70\,(\textcolor{Red}{$\downarrow 9.22$}) & 67.22\,(\textcolor{ForestGreen}{$\uparrow 4.30$}) \\
Claude 3.7 Sonnet (thinking) & 60.50 & 59.69\,(\textcolor{Red}{$\downarrow 0.81$}) & 64.16\,(\textcolor{ForestGreen}{$\uparrow 3.66$})& 51.52\,(\textcolor{Red}{$\downarrow 8.98$}) & 67.11\,(\textcolor{ForestGreen}{$\uparrow 6.61$}) \\
Gemini 2.5 Pro Preview& 58.65 & 57.99\,(\textcolor{Red}{$\downarrow 0.66$}) & 57.70\,(\textcolor{Red}{$\downarrow 0.95$})& 55.76\,(\textcolor{Red}{$\downarrow 2.89$}) & 65.07\,(\textcolor{ForestGreen}{$\uparrow 6.42$}) \\
GPT-5.1 & 58.33 & 61.47\,(\textcolor{ForestGreen}{$\uparrow 3.14$}) & 56.69\,(\textcolor{Red}{$\downarrow 1.64$})& 51.44\,(\textcolor{Red}{$\downarrow 6.89$}) & 63.69\,(\textcolor{ForestGreen}{$\uparrow 5.36$}) \\
OpenAI o4-mini-high & 59.72 & 61.68\,(\textcolor{ForestGreen}{$\uparrow 1.96$}) & 66.30\,(\textcolor{ForestGreen}{$\uparrow 6.58$})& 56.72\,(\textcolor{ForestGreen}{$\uparrow 3.00$}) & 66.38\,(\textcolor{ForestGreen}{$\uparrow 6.66$}) \\
Doubao-Seed-1.6 & 57.81 & 57.70\,(\textcolor{Red}{$\downarrow 0.11$}) & 56.31\,(\textcolor{Red}{$\downarrow 1.50$})& 54.84\,(\textcolor{Red}{$\downarrow 2.97$}) & 62.39\,(\textcolor{ForestGreen}{$\uparrow 4.58$}) \\
GPT-4o & 50.64 & 39.66\,(\textcolor{Red}{$\downarrow 10.98$}) & 55.10\,(\textcolor{ForestGreen}{$\uparrow 4.46$})& 56.99\,(\textcolor{ForestGreen}{$\uparrow 6.35$}) & 55.96\,(\textcolor{ForestGreen}{$\uparrow 5.32$}) \\
Grok 4.1 Fast & 52.09 & 51.39\,(\textcolor{Red}{$\downarrow 0.70$}) & 51.70\,(\textcolor{Red}{$\downarrow 0.39$})& 46.43\,(\textcolor{Red}{$\downarrow 5.66$}) & 58.35\,(\textcolor{ForestGreen}{$\uparrow 6.26$}) \\

\midrule
\noalign{\vspace{-0.5ex}}
\rowcolor{gray!8}\multicolumn{6}{c}
{\textbf{\textit{Open-Source MLLMs}}}\\
\midrule

GLM-4.5V & 59.12 & 58.81\,(\textcolor{Red}{$\downarrow 0.31$}) & 59.47\,(\textcolor{ForestGreen}{$\uparrow 0.35$})& 51.79\,(\textcolor{Red}{$\downarrow 7.33$}) & 64.77\,(\textcolor{ForestGreen}{$\uparrow 5.65$}) \\
Ministral 3 14B & 56.25 & 51.05\,(\textcolor{Red}{$\downarrow 5.20$}) & 54.36\,(\textcolor{Red}{$\downarrow 1.89$})& 42.80\,(\textcolor{Red}{$\downarrow 14.45$}) & 70.52\,(\textcolor{ForestGreen}{$\uparrow 14.27$}) \\
Ministral 3 8B & 55.19 & 46.20\,(\textcolor{Red}{$\downarrow 8.99$}) & 53.53\,(\textcolor{Red}{$\downarrow 1.66$})& 48.93\,(\textcolor{Red}{$\downarrow 6.26$}) & 68.88\,(\textcolor{ForestGreen}{$\uparrow 13.69$}) \\
Llama 4 Maverick & 56.03 & 53.44\,(\textcolor{Red}{$\downarrow 2.59$}) & 56.30\,(\textcolor{ForestGreen}{$\uparrow 0.27$})& 49.18\,(\textcolor{Red}{$\downarrow 6.85$}) & 58.41\,(\textcolor{ForestGreen}{$\uparrow 2.38$}) \\
Qwen3-VL-30B-A3B-Thinking & 53.99 & 50.56\,(\textcolor{Red}{$\downarrow 3.43$}) & 58.29\,(\textcolor{ForestGreen}{$\uparrow 4.30$})& 49.28\,(\textcolor{Red}{$\downarrow 4.71$}) & 62.62\,(\textcolor{ForestGreen}{$\uparrow 8.63$}) \\
InternVL3-78B & 49.36 & 44.98\,(\textcolor{Red}{$\downarrow 4.38$}) & 52.61\,(\textcolor{ForestGreen}{$\uparrow 3.25$})& 51.65\,(\textcolor{ForestGreen}{$\uparrow 2.29$}) & 55.47\,(\textcolor{ForestGreen}{$\uparrow 6.11$}) \\
Mistral Small 3.1& 49.36 & 29.56\,(\textcolor{Red}{$\downarrow 19.80$}) & 56.22\,(\textcolor{ForestGreen}{$\uparrow 6.86$})& 51.33\,(\textcolor{ForestGreen}{$\uparrow 1.97$}) & 62.07\,(\textcolor{ForestGreen}{$\uparrow 12.71$}) \\
Qwen3-VL-30B-A3B-Instruct & 49.50 & 40.45\,(\textcolor{Red}{$\downarrow 9.05$}) & 51.94\,(\textcolor{ForestGreen}{$\uparrow 2.44$})& 58.21\,(\textcolor{ForestGreen}{$\uparrow 8.71$}) & 52.48\,(\textcolor{ForestGreen}{$\uparrow 2.98$}) \\
Gemma 3 27B & 44.33 & 26.77\,(\textcolor{Red}{$\downarrow 17.56$}) & 49.76\,(\textcolor{ForestGreen}{$\uparrow 5.43$})& 61.29\,(\textcolor{ForestGreen}{$\uparrow 16.96$}) & 48.07\,(\textcolor{ForestGreen}{$\uparrow 3.74$}) \\

Qwen2.5-VL-72B & 44.38 & 
25.30\,(\textcolor{Red}{$\downarrow 19.08$}) & 
58.10\,(\textcolor{ForestGreen}{$\uparrow 13.72$})& 
47.27\,(\textcolor{ForestGreen}{$\uparrow 2.89$}) & 
64.52\,(\textcolor{ForestGreen}{$\uparrow 20.14$}) \\

LLaVA-v1.5-13B & 31.75 & 
20.15\,(\textcolor{Red}{$\downarrow 11.60$}) & 
46.81\,(\textcolor{ForestGreen}{$\uparrow 15.06$})& 
35.90\,(\textcolor{ForestGreen}{$\uparrow 4.15$}) & 
52.88\,(\textcolor{ForestGreen}{$\uparrow 20.13$}) \\

LLaVA-OneVision-7B & 31.49 & 
18.75\,(\textcolor{Red}{$\downarrow 12.74$}) & 
45.26\,(\textcolor{ForestGreen}{$\uparrow 13.77$})& 
34.34\,(\textcolor{ForestGreen}{$\uparrow 2.85$}) & 
51.15\,(\textcolor{ForestGreen}{$\uparrow 19.66$}) \\

\bottomrule
\end{tabular}
}
\caption{Performance breakdown for fine-grained staining images within the Panel-Level Fine-Grained Perception stage. Upward/downward arrows (\textcolor{ForestGreen}{$\uparrow$}/\textcolor{Red}{$\downarrow$}) indicate the absolute deviation from the model's corresponding Micro Avg. within this stage.}
\label{app:Tab.5}
\end{table*}

\begin{table*}[b]
\centering
\resizebox{\textwidth}{!}{
\begin{tabular}{l c c c c c c}
\toprule
\textbf{Model} & \textbf{Micro Avg.} & \textbf{OS} & \textbf{CE} & \textbf{AE} & \textbf{IS} &\textbf{P\&I} \\

\midrule
\noalign{\vspace{-0.5ex}}
\rowcolor{gray!8}\multicolumn{7}{c}
{\textbf{\textit{Proprietary MLLMs}}}\\
\midrule

Gemini 3 Pro Preview& 70.29 & 72.50\,(\textcolor{ForestGreen}{$\uparrow 2.21$}) & 72.79\,(\textcolor{ForestGreen}{$\uparrow 2.50$}) & 73.25\,(\textcolor{ForestGreen}{$\uparrow 2.96$}) & 65.18\,(\textcolor{Red}{$\downarrow 5.11$}) & 47.27\,(\textcolor{Red}{$\downarrow 23.02$}) \\
Claude 3.7 Sonnet (thinking) & 69.93 & 73.71\,(\textcolor{ForestGreen}{$\uparrow 3.78$}) & 67.57\,(\textcolor{Red}{$\downarrow 2.36$})& 74.25\,(\textcolor{ForestGreen}{$\uparrow 4.32$}) & 66.67\,(\textcolor{Red}{$\downarrow 3.26$}) & 54.90\,(\textcolor{Red}{$\downarrow 15.03$})\\
Gemini 2.5 Pro Preview& 68.24 & 71.13\,(\textcolor{ForestGreen}{$\uparrow 2.89$}) & 70.04\,(\textcolor{ForestGreen}{$\uparrow 1.80$}) & 71.43\,(\textcolor{ForestGreen}{$\uparrow 3.19$}) & 65.77\,(\textcolor{Red}{$\downarrow 2.47$}) & 37.25\,(\textcolor{Red}{$\downarrow 30.99$}) \\
GPT-5.1 & 67.23 & 73.76\,(\textcolor{ForestGreen}{$\uparrow 6.53$}) & 65.44\,(\textcolor{Red}{$\downarrow 1.79$})& 73.97\,(\textcolor{ForestGreen}{$\uparrow 6.74$}) & 57.89\,(\textcolor{Red}{$\downarrow 9.34$}) & 41.82\,(\textcolor{Red}{$\downarrow 25.41$}) \\
OpenAI o4-mini-high & 66.44 & 65.05\,(\textcolor{Red}{$\downarrow 1.39$}) & 64.71\,(\textcolor{Red}{$\downarrow 1.73$})& 72.00\,(\textcolor{ForestGreen}{$\uparrow 5.56$}) & 50.42\,(\textcolor{Red}{$\downarrow 16.02$}) & 45.28\,(\textcolor{Red}{$\downarrow 21.16$}) \\
Doubao-Seed-1.6 & 63.43 & 71.78\,(\textcolor{ForestGreen}{$\uparrow 8.35$}) & 61.40\,(\textcolor{Red}{$\downarrow 2.03$})& 65.84\,(\textcolor{ForestGreen}{$\uparrow 2.41$}) & 63.16\,(\textcolor{Red}{$\downarrow 0.27$}) & 32.73\,(\textcolor{Red}{$\downarrow 30.70$}) \\
GPT-4o & 60.73 & 68.66\,(\textcolor{ForestGreen}{$\uparrow 7.93$}) & 67.16\,(\textcolor{ForestGreen}{$\uparrow 6.43$})& 69.14\,(\textcolor{ForestGreen}{$\uparrow 8.41$}) & 48.67\,(\textcolor{Red}{$\downarrow 12.06$}) & 23.64\,(\textcolor{Red}{$\downarrow 37.09$}) \\
Grok 4.1 Fast & 57.79 & 61.88\,(\textcolor{ForestGreen}{$\uparrow 4.09$}) & 56.25\,(\textcolor{Red}{$\downarrow 1.54$})& 63.37\,(\textcolor{ForestGreen}{$\uparrow 5.58$}) & 53.51\,(\textcolor{Red}{$\downarrow 4.28$}) & 34.55\,(\textcolor{Red}{$\downarrow 23.24$}) \\
\midrule
\noalign{\vspace{-0.5ex}}
\rowcolor{gray!8}\multicolumn{7}{c}
{\textbf{\textit{Open-Source MLLMs}}}\\
\midrule
GLM-4.5V & 66.59 & 68.81\,(\textcolor{ForestGreen}{$\uparrow 2.22$}) & 68.75\,(\textcolor{ForestGreen}{$\uparrow 2.16$})& 71.19\,(\textcolor{ForestGreen}{$\uparrow 4.60$}) & 57.02\,(\textcolor{Red}{$\downarrow 9.57$}) & 47.27\,(\textcolor{Red}{$\downarrow 19.32$}) \\
Ministral 3 14B & 62.49 & 64.85\,(\textcolor{ForestGreen}{$\uparrow 2.36$}) & 67.28\,(\textcolor{ForestGreen}{$\uparrow 4.79$})& 64.20\,(\textcolor{ForestGreen}{$\uparrow 1.71$}) & 53.98\,(\textcolor{Red}{$\downarrow 8.50$}) & 40.00\,(\textcolor{Red}{$\downarrow 22.49$}) \\
Ministral 3 8B & 59.77 & 59.90\,(\textcolor{ForestGreen}{$\uparrow 0.13$}) & 61.40\,(\textcolor{ForestGreen}{$\uparrow 1.62$})& 69.42\,(\textcolor{ForestGreen}{$\uparrow 9.65$}) & 51.75\,(\textcolor{Red}{$\downarrow 8.02$}) & 25.45\,(\textcolor{Red}{$\downarrow 34.32$}) \\
Llama 4 Maverick & 67.01 & 71.43\,(\textcolor{ForestGreen}{$\uparrow 4.42$}) & 67.08\,(\textcolor{ForestGreen}{$\uparrow 0.07$})& 75.42\,(\textcolor{ForestGreen}{$\uparrow 8.41$}) & 70.97\,(\textcolor{ForestGreen}{$\uparrow 3.96$}) & 34.29\,(\textcolor{Red}{$\downarrow 33.72$}) \\
Qwen3-VL-30B-A3B-Thinking & 61.75 & 65.84\,(\textcolor{ForestGreen}{$\uparrow 4.09$}) & 63.97\,(\textcolor{ForestGreen}{$\uparrow 2.22$})& 63.49\,(\textcolor{ForestGreen}{$\uparrow 1.74$}) & 58.41\,(\textcolor{Red}{$\downarrow 3.34$}) & 33.96\,(\textcolor{Red}{$\downarrow 27.79$}) \\
InternVL3-78B & 59.80 & 61.89\,(\textcolor{ForestGreen}{$\uparrow 2.09$}) & 60.22\,(\textcolor{ForestGreen}{$\uparrow 0.42$})& 62.86\,(\textcolor{ForestGreen}{$\uparrow 3.06$}) & 59.70\,(\textcolor{Red}{$\downarrow 0.10$}) & 40.49\,(\textcolor{Red}{$\downarrow 19.31$}) \\
Mistral Small 3.1& 56.84 & 64.95\,(\textcolor{ForestGreen}{$\uparrow 8.11$}) & 49.61\,(\textcolor{Red}{$\downarrow 7.22$})& 63.76\,(\textcolor{ForestGreen}{$\uparrow 6.92$}) & 55.66\,(\textcolor{Red}{$\downarrow 1.18$}) & 35.19\,(\textcolor{Red}{$\downarrow 21.65$}) \\
Qwen3-VL-30B-A3B-Instruct & 57.00 & 56.44\,(\textcolor{Red}{$\downarrow 0.56$}) & 65.07\,(\textcolor{ForestGreen}{$\uparrow 8.07$})& 56.38\,(\textcolor{Red}{$\downarrow 0.62$}) & 52.63\,(\textcolor{Red}{$\downarrow 4.37$}) & 30.91\,(\textcolor{Red}{$\downarrow 26.09$}) \\
Gemma 3 27B & 55.95 & 63.35\,(\textcolor{ForestGreen}{$\uparrow 7.40$}) & 61.22\,(\textcolor{ForestGreen}{$\uparrow 5.27$})& 56.84\,(\textcolor{ForestGreen}{$\uparrow 0.89$}) & 43.64\,(\textcolor{Red}{$\downarrow 12.31$}) & 23.53\,(\textcolor{Red}{$\downarrow 32.42$}) \\

Qwen2.5-VL-72B & 59.95 & 
63.72\,(\textcolor{ForestGreen}{$\uparrow 3.77$}) & 
62.15\,(\textcolor{ForestGreen}{$\uparrow 2.20$})& 
61.50\,(\textcolor{ForestGreen}{$\uparrow 1.55$}) & 
54.08\,(\textcolor{Red}{$\downarrow 5.87$}) & 
35.20\,(\textcolor{Red}{$\downarrow 24.75$}) \\

LLaVA-v1.5-13B & 45.20 &
49.04\,(\textcolor{ForestGreen}{$\uparrow 3.84$}) & 
45.91\,(\textcolor{ForestGreen}{$\uparrow 0.71$})& 
48.26\,(\textcolor{ForestGreen}{$\uparrow 3.06$}) & 
42.80\,(\textcolor{Red}{$\downarrow 2.40$}) & 
31.67\,(\textcolor{Red}{$\downarrow 13.53$}) \\

LLaVA-OneVision-7B & 40.34 &
42.55\,(\textcolor{ForestGreen}{$\uparrow 2.21$}) & 
42.36\,(\textcolor{ForestGreen}{$\uparrow 2.02$})& 
43.64\,(\textcolor{ForestGreen}{$\uparrow 3.30$}) & 
35.65\,(\textcolor{Red}{$\downarrow 4.69$}) & 
26.32\,(\textcolor{Red}{$\downarrow 14.02$}) \\

\bottomrule
\end{tabular}
}

\caption{Performance breakdown across typical experimental paradigms within the Expert-Level Reasoning stage. Upward/downward arrows (\textcolor{ForestGreen}{$\uparrow$}/\textcolor{Red}{$\downarrow$}) indicate the absolute deviation from the model's corresponding Micro Avg. within this stage.}
\label{app:Tab.6}
\end{table*}

\clearpage
\onecolumn

\end{document}